\documentclass{article}

    \PassOptionsToPackage{square, comma, sort&compress, numbers}{natbib}

\usepackage[preprint]{neurips_2024}

\usepackage[utf8]{inputenc} 
\usepackage[T1]{fontenc}    
\usepackage{hyperref}       
\usepackage{url}            
\usepackage{booktabs}       
\usepackage{amsfonts}       
\usepackage{nicefrac}       
\usepackage{microtype}      
\usepackage{xcolor}         
\usepackage{ulem}
\usepackage{graphicx}
\usepackage{subcaption}
\usepackage{multirow}
\usepackage{listings}
\usepackage{wrapfig}
\usepackage[ruled,linesnumbered]{algorithm2e}
\usepackage{amsmath}
\usepackage{bbm}

\hypersetup{
  colorlinks=true
}

\definecolor{lightgray}{gray}{0.9}
\title{
\includegraphics[width=0.05\textwidth]{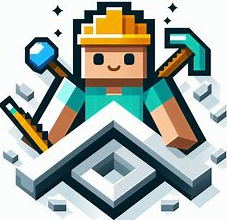}
Luban: Building Open-Ended Creative\\ Agents via Autonomous Embodied Verification}

\newcommand{\USTC}{University of Science and Technology of China}
\newcommand{\UCAS}{University of Chinese Academy of Sciences}
\newcommand{\SKLP}{State Key Lab of Processors, Institute of Computing Technology, CAS}
\newcommand{\CT}{Cambricon Technologies}
\newcommand{\SHIC}{Shanghai Innovation Center for Processor Technologies}
\newcommand{\ISCAS}{Intelligent Software Research Center, Institute of Software, CAS}
\newcommand{\ICT}{Institute of Computing Technology, CAS}

\author{
    Yuxuan~Guo\textsuperscript{1, 2, 3} \quad Shaohui~Peng\textsuperscript{6} \quad Jiaming~Guo\textsuperscript{2} \quad Di~Huang\textsuperscript{2} \quad Xishan~Zhang\textsuperscript{2, 3} \\
    \textbf{Rui~Zhang\textsuperscript{2} \quad Yifan~Hao\textsuperscript{2} \quad Ling~Li\textsuperscript{6} \quad Zikang~Tian\textsuperscript{2, 3, 4} \quad Mingju~Gao\textsuperscript{2, 3, 4}} \\
    \textbf{Yutai~Li\textsuperscript{2, 3, 4} \quad Yiming~Gan\textsuperscript{7} \quad Shuai~Liang\textsuperscript{7} \quad Zihao~Zhang\textsuperscript{2} \quad Zidong~Du\textsuperscript{2, 5}} \\
    \textbf{Qi~Guo\textsuperscript{2} \quad Xing~Hu\textsuperscript{2, 5~\footnotemark[1]} \quad Yunji~Chen\textsuperscript{2, 4~\thanks{Corresponding author.}}}\\ \\
    \textsuperscript{1}\USTC{}\\
    \textsuperscript{2}\SKLP{}\\
    \textsuperscript{3}\CT{} \quad \textsuperscript{4}\UCAS{}\\
    \textsuperscript{5}\SHIC{}\\
    \textsuperscript{6}\ISCAS{}\\
    \textsuperscript{7}\ICT{}\\ \\
    {\tt \small  gyx\_20170818@mail.ustc.edu.cn, \{huxing, cyj\}@ict.ac.cn} \\
}

\begin{document}

\maketitle

\begin{abstract}

Building open agents has always been the ultimate goal in AI research, and creative agents are the more enticing.
Existing LLM agents excel at long-horizon tasks with well-defined goals (e.g., `mine diamonds' in Minecraft).
However, they encounter difficulties on creative tasks with open goals and abstract criteria due to the inability to bridge the gap between them, thus lacking feedback for self-improvement in solving the task.
In this work, we introduce autonomous embodied verification techniques for agents to fill the gap, laying the groundwork for creative tasks.
Specifically, we propose the Luban agent target creative building tasks in Minecraft, which equips with two-level autonomous embodied verification inspired by human design practices: (1) visual verification of 3D structural speculates, which comes from agent synthesized CAD modeling programs; (2) pragmatic verification of the creation by generating and verifying environment-relevant functionality programs based on the abstract criteria.
Extensive multi-dimensional human studies and Elo ratings show that the Luban completes diverse creative building tasks in our proposed benchmark and outperforms other baselines ($33\%$ to $100\%$) in both visualization and pragmatism.
Additional demos on the real-world robotic arm show the creation potential of the Luban in the physical world.

\end{abstract}

\section{Introduction}
Developing open-ended agents capable of autonomously solving complex tasks \cite{saycan,codeaspolicy,instruct2act,innerm,Voyager,DEPS,hyvin} directed by high-level abstract instructions is the ultimate goal of artificial intelligence (AI) techniques.
Within this endeavor, creative tasks stand out as particularly enticing \cite{minedojo,Groot,pku-creative-agent}. 
Unlike conventional long-horizontal complex tasks, creative tasks do not have well-defined or easily automated success criteria \cite{minedojo}, thus stimulating the emergence of more advanced AI techniques with higher intellectual capabilities and the potential to tackle real-world problems.

\begin{figure}
    \centering
    \includegraphics[width=\linewidth]{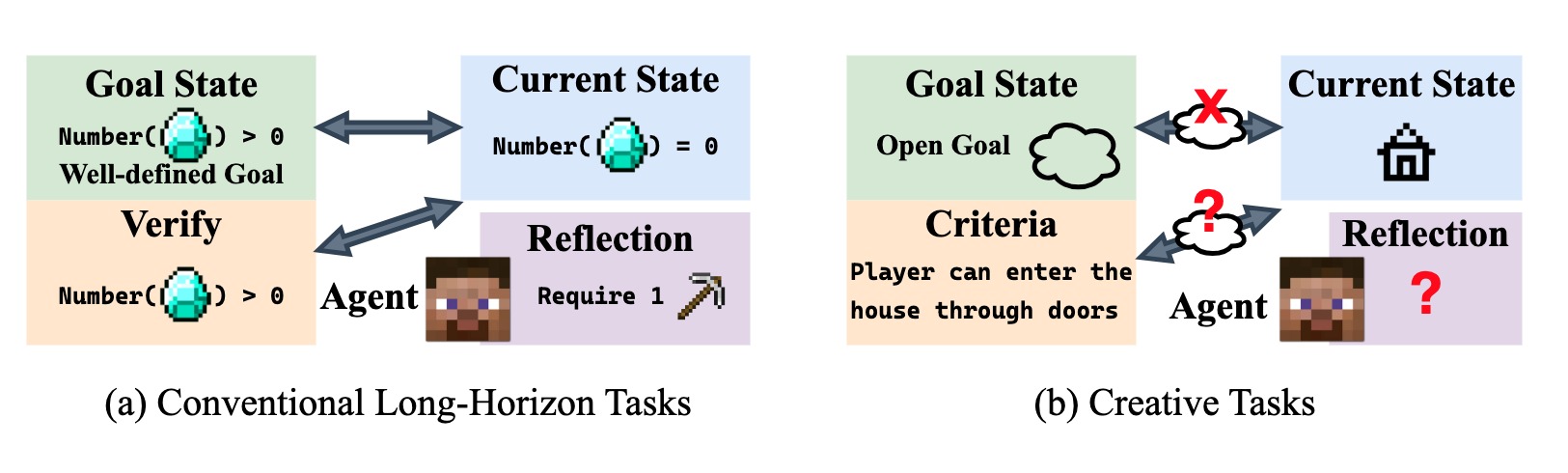}
    \caption{(a) Agents for Well-defined long-horizontal tasks v.s. (b) Luban agent for creative tasks.}
    \label{fig:motivation}
\end{figure}

Existing LLM agent techniques show promising progress in handling conventional long-horizon tasks with well-defined goals related to environment states.
For instance, as shown in Figure \ref{fig:motivation}, consider the classic `mining diamonds' task in Minecraft, where it is easy to verify by just checking the diamond number in the inventory.
Empowered by LLM's rich semantic knowledge and reasoning capabilities \cite{gpt-4,lamma2}, these agents \cite{Voyager,GITM,DEPS} {precisely} assess the distance between current states and the goals, then reflect and replan based on the assessment accordingly, eventually solving diamond mining tasks iteratively. 

However, when confronted with creative tasks, existing agents encounter a significant challenge: \textbf{the inability to verify or assess due to the absence of well-defined goals}, which is the prerequisite of reflection and re-planing.
For instance, the creative task of `building a house' in Minecraft lacks explicit goal definitions.
It is challenging for LLM agents to verify whether a house has been properly constructed and reflect based on the verification. 
For the assess criterion of \textit{`Player can enter the house through doors'}, there is a huge gap between high-level abstract descriptions (\textit{`enter through doors'}) and environment-relevant verification actions (\texttt{move}($(x_1, y_1, z_1) \to (x_2, y_2, z_2)$)). 
Such a gap impedes agents from accurately assessing their current states and formulating practical plans. 

To address this issue, we introduce a novel approach termed autonomous embodied verification techniques, aimed at empowering open-ended agents to \textbf{proficiently confirm high-level abstractions of assess criteria} in creative building tasks. 
This lays the groundwork for agents to autonomously tackle open-ended creative tasks without well-defined goals. 
We draw inspiration from human design practices that usually progressively design and verify from the visual appearance to functionality. 
Based on such inspiration, we propose a Luban agent, 
which begins with a building `something-like' phase, wherein we speculatively construct 3D structural objects based on CAD (Computer-Aided Design) program synthesis and perform visual verification on these objects. After passing the visual verification, it subsequently transits to the building `something-work' phase. Luban then generates environment-relevant functionality programs on these objects for pragmatic verification. With such visual and pragmatic verification, agents can summarize and reflect accordingly and iteratively complete open-ended creative tasks.

To evaluate the performance of Luban on open creative building tasks, we designed a benchmark containing 5 Minecraft building tasks with diverse visual and functional requirements.
Multi-dimensional extensive human studies show that the Luban agent successfully completes all open-ended creative building tasks, and the Elo ratings clearly show that buildings created by Luban outperform other baselines (33\%\textasciitilde100\%) in visualization and pragmatism.
Moreover, the pass rate of autonomously proposed embodied verification is consistent with human functionality assessment, demonstrating its effectiveness and necessity.
Finally, demos on the real-world robotic arm show the potential of the Luban agent to perform open-ended creative tasks in the physical world.

\section{Related Works} \label{sec:related-works}

\textbf{Minecraft Agents}.
The openness and authenticity of the Minecraft game make it an important test-bed for AI agents.
Most existing Minecraft agents focus on tasks with a long horizon and well-defined goals \cite{MCU}, such as collecting and crafting materials.
These agents can be further categorized into two branches: control-centric and planning-centric.
The control-centric agents \cite{VPT,Steve-1,Jarvis-1,Groot} trained on Minecraft gameplay demos collected from the Internet to build task policies based on low-level game controls (e.g., mouse and keyboard action).
The planning-centric agents \cite{Voyager,DEPS,GITM} focus on aligning high-level instructions with action primitives by utilizing LLM's reasoning capabilities and semantic knowledge to decompose instructions into plans.
These agents often come with carefully designed memory and reflection mechanisms to ensure they can learn useful skills and take advantage of environmental feedback.
Unlike the above works, we focus on building planning-centric creative agents that aim to autonomously verify the not well-defined goals of the creative tasks to ground creations (ensuring pragmatism) in the environment.
Compared with the pioneering attempt \cite{pku-creative-agent}, it did not involve any verification and feedback mechanisms, making it incompetent in grounding creations.

\textbf{3D Model Synthesis.}
Using computers to generate 3D models is a key research topic in computer graphics.
Recently, the synthesis of 3D models from given instructions (text or images) has attracted more and more attention from researchers \cite{3D-Gen-Survey1, 3D-Gen-Survey2, 3D-Gen-Survey3}.
The methods of 3D model synthesis can be divided into two categories.
One category methods synthesize 3D models directly (e.g., meshes \cite{3dgen}, point cloud \cite{point-e}, multi-view images \cite{dreamfusion} and voxels \cite{holo-diffusion}) rely on generative models \cite{VAE, GAN, NormalizedFlow, Transformer, DDPM} and neural representations \cite{nerf,3d-gaussian}.
Another category of methods relies on the existing Computer-Aided-Design (CAD) modeling software (e.g., Blender \cite{Blender} and FreeCAD \cite{FreeCAD}) to first synthesize the operations and parameters of the modeling process (i.e., programs) and then execute them to get the 3D model.
This line of work includes training-based methods \cite{Hierarchical-CAD, CADParser, Free2CAD} and LLM in-context learning-based method \cite{3D-GPT} that emerged recently.
The models synthesized using the former category of methods typically exhibit rich textures and details but lack complete controllability and accurate dimensions, whereas those synthesized using the latter demonstrate the opposite.
In this work, generating accurate 3D models is crucial to the Luban agent's planning and visual verification, so we synthesize 3D models by prompting LLM to synthesize programs based on the CAD modeling library we provided.
Compared with \cite{3D-GPT}, we consider CAD modeling to rely on a small number of low-level (i.e., sketch-extrude-based) rather than high-level (i.e., pre-defined objects) APIs, which allow the creation of diverse 3D models via using API combinations and adding natural language annotations.
Please refer to Sec. \ref{sec:method-1} and Appendix \ref{app:implementation} for more details.

\section{Problem Definition}
\label{sec:problem-defination}



\textbf{Minecraft Environment.}
We formalize the Minecraft environment as a Partially Observable Markov Decision Process (POMDP) without the reward function $P=(S, A, T, \Omega, O)$, where $S$ is the state space, $A$ is the action space, $T$ is the transition dynamics, $\Omega$ is the observation space (i.e., game images), and $O$ is the set of conditional observation probabilities.
The action space $A$ consists of pre-defined action primitives, such as \texttt{move}, \texttt{place\_block}, and \texttt{dig\_block}, which return a binary value to reveal action status (success or failure).

\textbf{Minecraft Agent for Open-ended Creative Building Tasks.}
The open-ended creative building tasks can be formalized as an Instruction Following (IF) problem, where the instruction $I$ consists of two parts:
(1) \textbf{Text}, including Natural Language (NL) building description, functional requirements, and building suggestions;
(2) \textbf{Images}, including multi-view images of a general example building that aligns with the building description.
The agent takes the instruction $I$ as input and performs a sequence of actions $(a_1, a_2, \dots), a_i \in A$ to build the building in the environment and ensures it meets the functional requirements (i.e., grounding creations in the environment by ensuring the pragmatism).
For example, when the instruction involves `build a bridge to cross a river', the agent should build a bridge-like structure in the environment and ensure it is walkable across the river.

\section{Method}

In this section, we introduce the Luban agent, which can complete open-ended creative building tasks pragmatically in the open world with the help of the two-level autonomous embodied verification:
(1) 3D structural speculation stage with the visual verification (Sec. \ref{sec:method-1});
(2) Construction stage with the pragmatic verification (Sec. \ref{sec:method-2}).

\begin{figure}
    \centering
    \includegraphics[width=\linewidth]{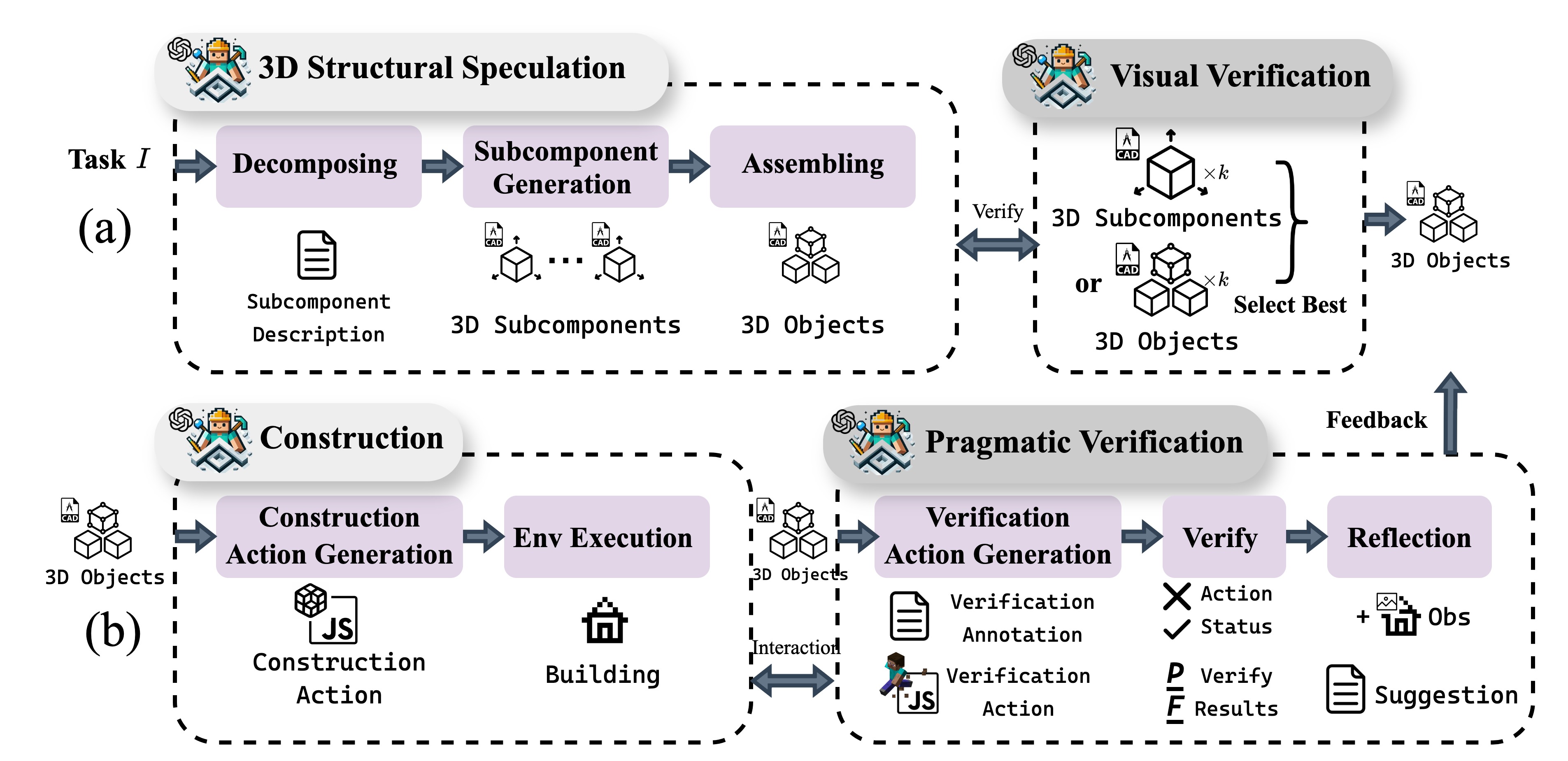}
    \caption{The diagram of Luban agent.
        (a)
        The \textbf{3D structural speculation} stage uses VLM to synthesize Instructions $I$ into a CAD program representing the building 3D objects, which further includes decomposing, subcomponents generation, and assembling.
        The \textbf{visual verification} evaluates the quality of buildings through the appearance results of the CAD program construction.
        (b)
        The \textbf{construction} stage uses VLM to synthesize the building's 3D object program into executable construction actions to get the building in the environment.
        The \textbf{pragmatic verification} evaluates the building 3D object's pragmatism by generating environment-relevant functionality annotations and action verify programs.
    }
    \label{fig:method}
\end{figure}

\subsection{3D Structural Speculation stage with Visual Verification}\label{sec:method-1}

The goal of the 3D structural speculation stage is to design the building based on the open-end creative building instruction $I$.
Due to the large goal space of the creative building tasks, it is necessary to introduce verifications in the 3D structural speculation stage that filter out the open but inappropriate designs (e.g., designs leading to semantic-less or incomplete building) to reduce the space.
We introduce visual verification in the 3D structural speculation stage by exploiting VLM's visual understanding capabilities, thus requiring the generation of visual representations.
Consider existing deep-learning-based visual representations synthesize techniques are inaccurate and uncontrollable (more discussion in Sec \ref{sec:related-works}),
we turn to synthesize parametrically modeled 3D models
(i.e., synthesizing Python CAD programs based on a Python CAD library \footnote{The Python CAD library used in our work is simplified and encapsulated based on the CadQuery project, which has a small number of low-level sketch-extrude-based APIs for parametric 3D modeling. Please refer to the Appendix \ref{app:implementation} for more details. The CadQuery project's link, \url{https://github.com/CadQuery/cadquery}.}) in the 3D structural speculation stage.
Figure \ref{fig:method} (a) shows the 3D structural speculation stage and visual verification.

\textbf{3D structural speculation.}
The 3D structural speculation stage can be formalized as $I \stackrel{\texttt{prompt}}{\longrightarrow} P^{B}$, where $P^{B}$ is a Python CAD program representing the precise 3D shape of the whole building.
To fully exploit VLM's 3D structural speculation and reasoning capabilities and consider the conventions of parametric CAD modeling, the 3D structural speculation stage is further divided into three sub-stages as shown in Figure \ref{fig:method} (a), including decomposing, subcomponent generation, and assembling.
(1) \textbf{Decomposing.} The VLM takes $I$ and necessary prompts as input and outputs a subcomponent description set $S=\{s_1, s_2, \dots\}$ that make up the building represented by $I$, expressed as $I \stackrel{\texttt{prompt}}{\longrightarrow} S$.
Each subcomponent $s_i$ in $S$ is represented in natural language and contains semantic, size, and position information.
(2) \textbf{Subcomponent Generation.}
The sub-stage aims at synthesizing natural language subcomponents into 3D subcomponents represented by Python CAD modeling programs $P^S$, expressed as $S \stackrel{\texttt{prompt}}{\longrightarrow} P^S$.
The VLM first plans to determine the precise size and appearance information of each subcomponent.
Then, it in-context learns the documents and few-shot examples of the CAD library we provide to synthesize the 3D subcomponents program.
(3) \textbf{Assembling.}
The VLM assembles the 3D subcomponents $P^S$ to the building 3D object by reasoning and setting each subcomponent's position and orientation via synthesizing building 3D object program $P^B$, expressed as $P^S \stackrel{\texttt{prompt}}{\longrightarrow} P^B$.

\textbf{Visual Verification.}
Visual verification aims to filter out the best from multiple modeling programs (i.e., 3D subcomponents and object), expressed as $I, (P^S_1, \dots, P^S_k) \stackrel{\texttt{prompt}}{\longrightarrow} i, 1 \le i \le k$ (or $(P^B_1, \dots, P^B_k)$).
Specifically, in the subcomponent generation and assembling sub-stages, we sample $k$ Python CAD modeling programs ($P^S$ or $P^B$) generated by VLM and execute them to get corresponding 3D model multi-view images.
Subsequently, the $k$ multi-view images are prompted to VLM to evaluate the consistency of the images and the instruction $I$.
The best program returned by VLM is selected, with the option to resample if no best program is found.

\subsection{Construction stage with the Pragmatic Verification} \label{sec:method-2}
\textbf{Construction.}
The construction stage aims to construct the building $B$ in the environment based on the building 3D object program $P^B$ from the 3D structural speculation stage, expressed as $P^B \stackrel{\texttt{prompt, execute}}{\longrightarrow} B$.
Specifically, the building 3D object program $P^B$ is first exported to the environment-level (i.e., Minecraft) coordinates.
Then, the VLM is prompted with these coordinates and the available action primitives to synthesize the action sequence $A^C = (a^C_1, a^C_2, \dots)$ (i.e., the JavaScript programs) for constructing the building.
Finally, the construction action sequence $A^C$ is executed in the environment to construct the building $B$.

\textbf{Pragmatic Verification.}
The pragmatic verification aims to reason well-defined functionality from the abstract criteria in task instruction $I$, and further verify corresponding pragmatism of the constructed building $B$ to get suggestions $I^+$ for improving the next round creation, expressed as $B \stackrel{\texttt{prompt, interact}}{\longrightarrow} I^+$. 
The Luban's pragmatic verification can be divided into three sub-stages, as shown in Figure \ref{fig:method}(b), including verification action generation, verify, and reflection.
(1) \textbf{Verification action generation.}
Based on the instruction $I$, the agent generates and attaches the natural language verification annotations on the subcomponents of the building 3D object $P^B$ to generate environment-relevant functionality programs.
The environment-relevant functionality programs are further synthesized into embodied verification actions (i.e., the JavaScript programs) with binary status (action success or not) by the VLM, i.e., $A^P = (a^P_1, a^P_2, \dots), a^P_i: B \to \{0, 1\}$.
(2) \textbf{Verify.}
The verification actions are further executed in the environment to interact with building $B$ to collect the action status.
By analyzing the action status, the agent verifies the building pragmatism and outputs verification results.
These two sub-stages are expressed as $I, P^S \stackrel{\texttt{prompt, execute}}{\longrightarrow} \{0, 1\}^n$.
(3) \textbf{Reflection.}
The verification results, together with instructions $I$ and image observation $o$, are prompted to VLM for further conducting semantic level check and reflection to obtain suggestions $I^+$ for the next iteration, expressed as $I, o, \{0, 1\}^n \stackrel{\texttt{prompt}}{\longrightarrow} I^+ $.

\section{Experiments}\label{sec:exp}
\begin{figure}
    \centering
    \includegraphics[width=\linewidth]{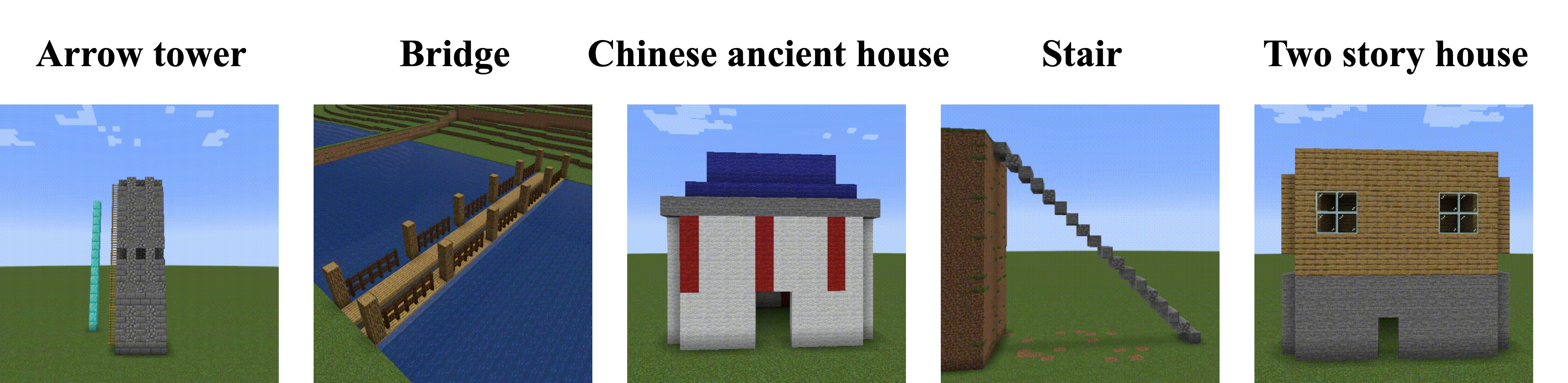}
    \caption{
        The showcases of Luban's creation on all tasks.
    }
    \label{fig:showcase-main}
\end{figure}

In this section, we first introduce the experimental settings (benchmark, baselines, and metrics) in Sec. \ref{sec:exp-1},
then demonstrate Luban's superiority in creation pragmatism and human preference compared to other method baselines and the quality of pragmatic verification in Sec. \ref{sec:exp-2},
further, show the effectiveness and necessity of Luban's two-level verifications through ablation studies in Sec. \ref{sec:exp-3},
and finally, discuss the real-world application potential of the Luban in Sec. \ref{sec:exp-4}.

\subsection{Exprimental Settings} \label{sec:exp-1}

\textbf{Benchmarking Open-ended Creative Building Tasks.} 
We design a benchmark to test the agent's ability to complete open-ended creative building tasks pragmatically.
The benchmark contains 5 tasks (i.e., \texttt{arrow-tower}, \texttt{bridge}, \texttt{chinese-ancient-house}, \texttt{stair}, and \texttt{two-story-house}) with diverse structural and functional requirements. 
Each task instruction consists of the text and multi-view images \footnote{Given the building descriptions, we use the text-to-3D service of \url{https://meshy.ai} to generate general 3D models and capture multi-view images.}. as illustrated in Sec. \ref{sec:problem-defination}.
Take the \texttt{bridge} task as an example. The task requires the agent to build a plank bridge similar to the one in multi-view images.
The functional requirements of the \texttt{bridge} task are two-fold: Environmental level, the bridge needs to cross the river; Building level, the bridge should be walkable for players, and the bridge's handrails should prevent players from falling off.
Please refer to Appendix \ref{app:benchmark} for more details about the benchmark and the comparison with other benchmarks.

\textbf{Baselines.}
We implement the \textbf{Luban} agent with \texttt{gpt-4-vision-preview}.
For simplicity, we assume that the action primitives used for construction (i.e., \texttt{place\_block} and \texttt{dig\_block}) and the building's position are oracles.
All action primitives are implemented with Mineflayer \cite{mineflayer} Javascript APIs and also adopted by the following baselines for the sake of fairness.
To demonstrate the superiority of the Luban agent, we compare it with the following plan-centric Minecraft agent baselines: 
(a) Voyager agent \cite{Voyager}, an LLM-based agent target on exploring the Minecraft world to follow instructions, which has 2 variants, \texttt{gpt-3.5-turbo} based \textbf{Voyager35} and \texttt{gpt-4} based \textbf{Voyager4};
(b) \textbf{Creative} agent \cite{pku-creative-agent}, a \texttt{gpt-4-vision-preview} based agent targets creative building tasks without any feedback from the environment.
To demonstrate the effectiveness and necessity of the Luban agent's two-level verification, we consider the following baseline of ablation settings:
(a) \textbf{Luban w/o pv}, Luban agent without pragmatic verification, equivalent to plan and construction without any environmental feedback;
(b) \textbf{Luban w/o vv}, Luban agent without visual verification by replacing the VLM visual verification with a random choice;
(c) \textbf{Luban w/o vvpv}, Luban agent without both verification.

\textbf{Metrics.}
We run all the above baselines to obtain 3 seed building results (multi-view video) for each task.
The 3 metrics are listed as follows:
(1) \textbf{Quality rating}, similar to \cite{pku-creative-agent}, each result is rated (ranging from 1 to 5) from 5 dimensions: \textbf{Appearance (\texttt{AP}), Complexity (\texttt{CO}), Aesthetics (\texttt{AE}), Building-level Functional (\texttt{FB}), and Environmental-level Functional (\texttt{FE}).}
The action verification of pragmatic verification (in Sec. \ref{sec:method-2}) mainly covers the \texttt{FB}, and the VLM semantic check followed by the action verification mainly covers the \texttt{FE}.
(2) \textbf{One-to-one comparison}, the result pairs from the same tasks and different baselines are evaluated by selecting the winner. The winning rates of baselines are further used to compute the Elo \cite{elo} rating for a comprehensive comparison.
(3) \textbf{Pass rate of Luban's pragmatic verification}, we migrate the pragmatic verification actions autonomously proposed by the Luban agent to other baselines (by modifying some parameters of the action primitives, e.g., start and end position of \texttt{move}) and execute the actions to calculate the pass rates for evaluation.
Considering the openness of the task results, we launch human studies on metrics (1) and (2). 
Please refer to Appendix \ref{app:human-study} for more details.

\subsection{Main Results} \label{sec:exp-2}

\begin{figure}
    \centering
    \includegraphics[width=\linewidth]{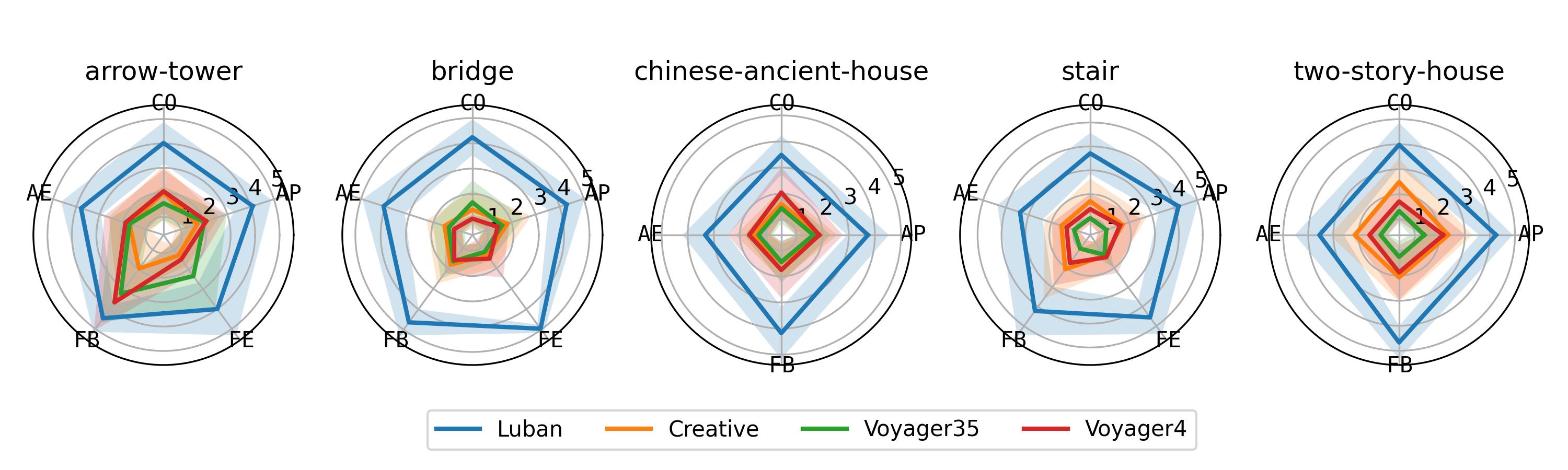}
    \caption{
        The polar chart of multi-dimensional quality rating of creations from Luban and other method baselines. The results are grouped by tasks and averaged across all seeds and human evaluators with a 1-sigma bar
    }
    \label{fig:h1-main}
\end{figure}

In this section, we comprehensively compare Luban and other method baselines on the 3 metrics: quality rating, one-to-one comparison, and pragmatic verification pass rate.
We draw 3 corresponding conclusions as follows: 

\textbf{Luban's creations outperform other method baselines and are pragmatic in the environment.}
As the quality ratings shown in the polar chart of Figure \ref{fig:h1-main}, on all 5 tasks, Luban's quality ratings on 3 non-functional dimensions significantly exceeded the baselines: \texttt{AE} rating increased by $1.42$ to $2.93$, \texttt{CO} rating increased by $1.44$ to $3.22$, and \texttt{AP} rating increased by $1.84$ to $3.18$.
Further, Luban also receives the highest ratings in the two functional dimensions, \texttt{FB} and \texttt{FE}, on all 5 tasks (average rating $4.44$ and $4.50$ correspondingly, near full rating $5$) and significantly outperforms the baselines of other methods (rating increased by $0.80$ to $3.42$ and $1.67$ to $3.76$ correspondingly).
The quality rating results directly demonstrate this conclusion.
Other method baselines cannot generate complicated creations and get feedback from the environment, thus resulting in low-level ratings.
We showcase Luban's creation in Figure \ref{fig:showcase-main} to facilitate a more intuitive understanding.
More showcases of other method baselines are shown in Figure \ref{app:fig:showcase-vs-other} in Appendix \ref{app:add-exp-show}.

\textbf{Luban's creation is more consistent with human preferences than other method baselines.}
As listed in Table \ref{table:h2-main}, on all 5 tasks, Luban achieves $\sim 100\%$ one-to-one winning rate compared with other method baselines.
The Elo rating across tasks provides a more comprehensive perspective to reflect the gap between baselines, where Luban outperforms the second baseline $\sim 500$ scores, directly supporting the conclusion.

\textbf{The pass rate of Luban's pragmatic verification reveals the degree of creation pragmatism.}
As the pass rates listed in Table \ref{table:h2-pv-passrate} left, Luban achieves $100\%$ verification pass rate on all 5 tasks after rounds of iteration and autonomous verification. In contrast, the pass rate of other method baselines remains low.
We observe the pass rates exhibit similar trending to the quality rating \texttt{FB}, and we statistically reveal it by computing the Spearman correlations, as listed in Table \ref{table:h2-pv-passrate} right.
The strong positive correlation (all $\rho > 0.6$ and \emph{p-value} $< 0.05$) indicates that the pragmatic verification pass rate aligns with the human evaluator. 
Thus, the pass rate reveals the degree of creation pragmatism and can measure creative building tasks.

\begin{table}
    \centering
    \caption{
        The winning rate ($\%$) of one-to-one comparison between Luban and other method baselines and Elo ratings across tasks.
    }
    \label{table:h2-main}
    \begin{tabular}{c|cccc}
        \toprule
        Task ID & Luban & Creative & Voyager35 & Voyager4 \\
        \midrule
        \texttt{arrow-tower} & \textbf{100.00} & 20.00 & 44.44 & 35.56 \\
        \texttt{bridge} & \textbf{100.00} & 55.56 & 1.39 & 43.06 \\
        \texttt{chinese-ancient-house} & \textbf{99.26} & 15.56 & 30.37 & 54.81 \\
        \texttt{stair} & \textbf{97.78} & 19.26 & 38.52 & 44.44 \\
        \texttt{two-story-house} & \textbf{98.52} & 62.22 & 6.67 & 32.59 \\
        \midrule
        Elo rating across tasks & \textbf{2095.83} & 1572.22 & 1053.55 & 1278.40 \\
        \bottomrule
    \end{tabular} 
\end{table}

\begin{table}
    \centering
    \caption{
        (\emph{Left}) The average pragmatic verification pass rate ($\%$) across seeds of Luban and other method baselines. (\emph{Right}) The Spearman correlation ($\rho$ and \emph{p-value}) between the pass rate and quality ratings \texttt{FB}.
    }
    \label{table:h2-pv-passrate}
    \begin{tabular}{c|cccc|c}
        \toprule
        Task ID & Luban & Creative & Voyager35 & Voyager4 & $\rho\ (p)$ \\
        \midrule
        \texttt{arrow-tower} & \textbf{100.00} & 33.33 & 66.67 & 100.00 & 0.76 (0.00)\\
        \texttt{bridge} & \textbf{100.00} & 22.22 & 22.22 & 33.33  & 0.89 (0.00) \\
        \texttt{chinese-ancient-house} & \textbf{100.00} & 0.00 & 0.00 & 0.00 & 0.75 (0.00) \\
        \texttt{stair} & \textbf{100.00} & 33.33 & 0.00 & 33.33 & 0.63 (0.03) \\
        \texttt{two-story-house} & \textbf{100.00} & 33.33 & 0.00 & 0.00 & 0.67 (0.02)\\
        \bottomrule
    \end{tabular} 
\end{table}

\subsection{Ablation Study} \label{sec:exp-3}

\begin{figure}
    \centering
    \includegraphics[width=\linewidth]{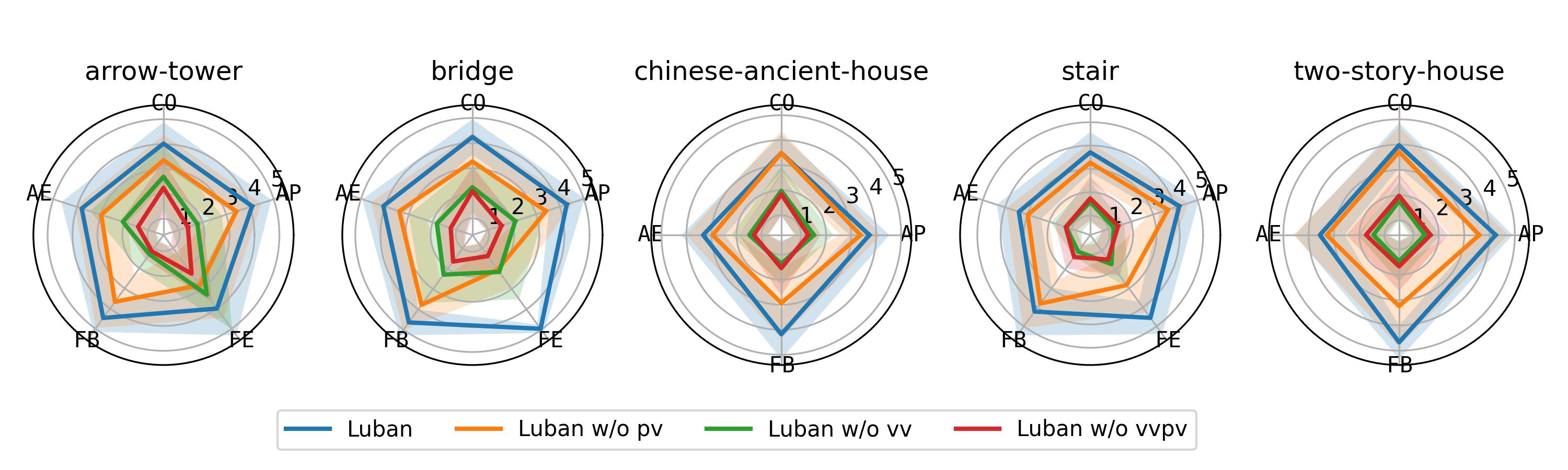}
    \caption{
        The polar chart of multi-dimensional quality rating of creations from Luban and ablation baselines. The results are grouped by tasks and averaged across all seeds and human evaluators with a 1-sigma bar.
        }
    \label{fig:h1-abl}
\end{figure}

In this section, we ablate Luban's visual and pragmatic verifications and draw 3 conclusions as follows: 

\textbf{Visual verification improves the basic quality of the creations.}
As the quality ratings shown in the polar chart of Figure \ref{fig:h1-abl}, on all 5 tasks, the quality ratings of baselines with visual verification (`Luban' and `Luban w/o pv') significantly outperform those without (`Luban w/o vv' and `Luban w/o vvpv') on both functional and non-functional dimensions (rating increasing from $0.69$ to $3.20$).
The one-to-one winning rates and Elo ratings in Table \ref{table:h3-abl} also exhibit similar trendings, in which `Luban' and `Luban w/o pv' are also significantly higher than `Luban w/o vv' and `Luban w/o vvpv'.
These results directly support the conclusion, and the creation quality degraded to GPT4 levels without visual verification.
The visual verification works because it filters out inappropriate Python CAD modeling programs to reduce the errors in subcomponent generation (e.g., missed or wrongly designed subcomponents) and assembling (e.g., incorrect subcomponent's position and orientation) sub-stages by reviewing multiple programs and selecting the best.

\textbf{Pragmatic verification is effective and necessary for the creation's pragmatism.}
As the two functional dimension ratings (\texttt{FB} and \texttt{FE}) shown in the polar chart of Figure \ref{fig:h1-abl}, on all 5 tasks, Luban outperforms `Luban w/o pv' baseline, ranging from $0.42$ to $1.43$ and $1.16$ to $2.91$ correspondingly.
Moreover, as the pragmatic verification pass rates listed in Table \ref{table:pv-abl}, `Luban w/o pv' does not reach $100\%$ pass on all tasks.
The differences between the above functional dimension ratings and verification pass rate directly demonstrate this conclusion.
The pragmatic verification works because it generates purposefully embodied actions to collect the information of creations for feedback to improve the creation's pragmatism stably. 
In contrast, those without pragmatic verification can rely solely on VLM output's randomness to make pragmatic creations occasionally.
Additionally, we notice that the verification pass rates of `Luban w/o pv' on tasks \texttt{bridge} and \texttt{stair} are 100\%, which may be attributed to the agreement of these Minecraft buildings and the VLM's semantic knowledge.

\textbf{Visual verification is the prerequisite for pragmatic verification.}
We access the pragmatic verification gains on baselines with and without visual verification by computing the two functional ratings (\texttt{FB} and \texttt{FE}) differences in Figure \ref{fig:h1-abl}, i.e., $\texttt{gain}(\text{Luban w/o vvpv} \to \text{Luban w/o vv}) = [-0.29, 1.04]$ and $\texttt{gain}(\text{Luban w/o pv} \to \text{Luban}) =  [0.42, 2.91]$.
The results show that larger pragmatic verification gains occurred in the baseline group with visual verification, which supports the conclusion.
This is because pragmatic verification means little when the building quality is extremely low, e.g., there is no point in verifying that the door is passable when the house is assembled incorrectly.
The results listed in Table \ref{table:pv-abl} support the reason, in which the two ablation baselines without visual verification (i.e., Luban w/o vv' and Luban w/o vvpv')'s pragmatic verification pass rate is no longer significantly correlated to the human functionality ratings (the p-values $>0.05$ on all 5 tasks).
More intuitive showcases of the ablation baselines can be found in Figure \ref{app:fig:showcase-vs-ablation} of Appendix \ref{app:add-exp-show}.

\begin{table}
    \centering
    \caption{
        The winning rate ($\%$) of one-to-one comparison between Luban and ablation baselines and Elo ratings across tasks.
    }
    \label{table:h3-abl}
    \begin{tabular}{c|cccc}
        \toprule
        Task ID & Luban & Luban w/o pv & Luban w/o vv & Luban w/o vvpv \\
        \midrule
        \texttt{arrow-tower} & \textbf{98.41} & 53.97 & 42.86 & 4.76 \\
        \texttt{bridge} & \textbf{98.41} & 65.08 & 35.71 & 0.79 \\
        \texttt{chinese-ancient-house} & \textbf{83.33} & 80.95 & 11.11 & 24.60 \\
        \texttt{stair} & 82.54 & \textbf{84.13} & 29.37 & 3.97 \\
        \texttt{two-story-house} & \textbf{86.51} & 76.98 & 17.46 & 19.05 \\
        \midrule
        Elo across tasks & \textbf{1979.48} & 1753.42 & 1128.74 & 1138.36 \\
        \bottomrule
    \end{tabular} 
\end{table}

\begin{table}
    \centering
    \caption{
        (\emph{Left}) The average pragmatic verification pass rate ($\%$) across seeds of Luban and ablation baselines. (\emph{Right}) The Spearman correlation ($\rho$ and \emph{p-value}) between the pass rate of (`Luban w/o vv', `Luban w/o vvpv') and quality ratings \texttt{FB}. The correlation item of \texttt{arrow-tower} task is `n/a' due to the constant pass rate.
    }
    \label{table:pv-abl}
    \begin{tabular}{c|ccc|c}
        \toprule
        Task ID & Luban w/o pv & Luban w/o vv & Luban w/o vvpv & $\rho\ (p)$ \\
        \midrule
        \texttt{arrow-tower} & 66.67 & 0.00 & 0.00 & n/a (n/a) \\
        \texttt{bridge} & 100.00 & 55.56 & 55.56 & 0.53 (0.28) \\
        \texttt{chinese-ancient-house} & 33.33 & 33.33 & 66.67 & 0.60 (0.21) \\
        \texttt{stair} & 100.00 & 0.00 & 25.00 & 0.77 (0.07) \\
        \texttt{two-story-house}& 33.33 & 0.00 & 33.33 & -$0.42$ (0.41)\\
        \bottomrule
    \end{tabular}
    
\end{table}

\subsection{Potential in Real-World Embodied Creative Tasks} \label{sec:exp-4}

\begin{figure}
    \centering
    \includegraphics[width=\linewidth]{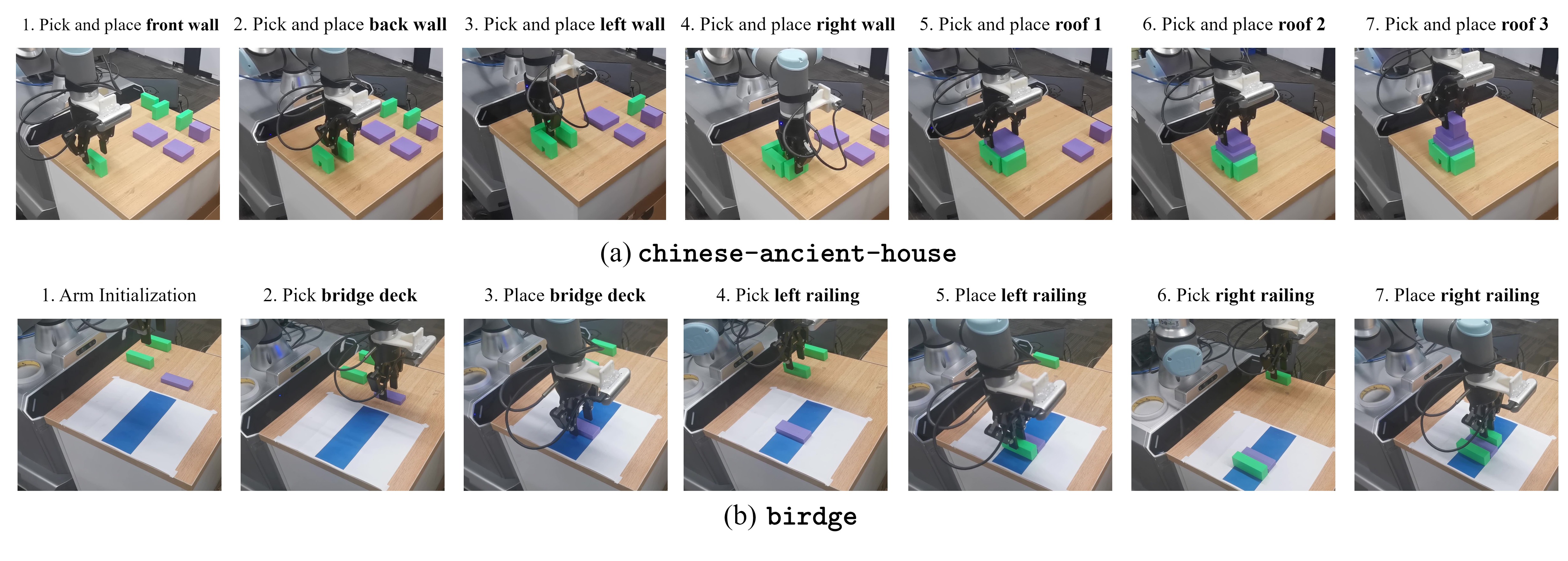}
    \caption{
        The robotic demos of task \texttt{chinese-ancient-house} and \texttt{bridge}.
    }
    \label{fig:5.4-demo}
\end{figure}

Luban has potential in real-world open-ended creative tasks rather than being limited to Minecraft-like virtual worlds, which owes to the general of Luban's planning framework (i.e., CAD modeling and visual verification) and feedback mechanism (i.e., propose actions to verify not well-defined goals).
We demonstrate this by providing demos on two tasks (\texttt{chinese-ancient-house} and \texttt{bridge}) of Luban on real-world embodied robotic arm environment, as shown in Figure \ref{fig:5.4-demo}.
Specifically, we first 3D print the model of subcomponents and then use the subcomponent coordinates given by the assembling sub-stage to drive the pick-place API-based embodied robotic arm to perform creative building tasks.
The final assembly result demonstrates the conclusion.

\section{Conclusion}

In this work, we propose Luban, an agent capable of open-ended creative building tasks in Minecraft, powered by the two-level autonomous embodied visual and pragmatic verifications.
Extensive human studies demonstrate that Luban's creations have higher quality (especially functional pragmatism) in multiple dimensions and are more preferred by humans than the other method baselines.
Furthermore, Luban also shows the potential of Luban in real-world creative tasks through demos we provided on embodied robotic arms environment. Our work may inspire the following directions:
(1) Develop libraries that represent the 3D physical world to bridge VLM and the physical world, thereby facilitating the emergence of embodied agents with spatial intelligence.
(2) Extend Luban's pragmatic verification to obtain feedback in the real world, thereby building a closed-loop, open creative agent grounding in the real world.

\section{Limitations} \label{sec:limitations}
 
We summarize our limitations as follows:
(1) Due to the lack of a memory mechanism, Luban cannot utilize shared knowledge between multiple tasks (e.g., universal design guidelines) and thus cannot learn from the environment continuously;
(2) The expensive access costs and limited capabilities of advanced VLM (i.e., GPT-4V) prevent Luban from generating more refined 3D structure inference.




\bibliography{ref}

\begin{thebibliography}{10}

\bibitem{saycan}
Michael Ahn, Anthony Brohan, Noah Brown, Yevgen Chebotar, Omar Cortes, Byron David, Chelsea Finn, Chuyuan Fu, Keerthana Gopalakrishnan, Karol Hausman, Alex Herzog, Daniel Ho, Jasmine Hsu, Julian Ibarz, Brian Ichter, Alex Irpan, Eric Jang, Rosario~Jauregui Ruano, Kyle Jeffrey, Sally Jesmonth, Nikhil Joshi, Ryan Julian, Dmitry Kalashnikov, Yuheng Kuang, Kuang-Huei Lee, Sergey Levine, Yao Lu, Linda Luu, Carolina Parada, Peter Pastor, Jornell Quiambao, Kanishka Rao, Jarek Rettinghouse, Diego Reyes, Pierre Sermanet, Nicolas Sievers, Clayton Tan, Alexander Toshev, Vincent Vanhoucke, Fei Xia, Ted Xiao, Peng Xu, Sichun Xu, Mengyuan Yan, and Andy Zeng.
\newblock Do as i can and not as i say: Grounding language in robotic affordances.
\newblock In {\em arXiv preprint arXiv:2204.01691}, 2022.

\bibitem{VPT}
Bowen Baker, Ilge Akkaya, Peter Zhokov, Joost Huizinga, Jie Tang, Adrien Ecoffet, Brandon Houghton, Raul Sampedro, and Jeff Clune.
\newblock Video pretraining (vpt): Learning to act by watching unlabeled online videos.
\newblock In Alice~H. Oh, Alekh Agarwal, Danielle Belgrave, and Kyunghyun Cho, editors, {\em Advances in Neural Information Processing Systems}, 2022.

\bibitem{Groot}
Shaofei Cai, Bowei Zhang, Zihao Wang, Xiaojian Ma, Anji Liu, and Yitao Liang.
\newblock Groot: Learning to follow instructions by watching gameplay videos.
\newblock In {\em The Twelfth International Conference on Learning Representations}, 2024.

\bibitem{Blender}
Blender~Online Community.
\newblock Blender - a 3d modelling and rendering package, 2018.

\bibitem{elo}
A.E. Elo.
\newblock {\em The USCF Rating System: Its Development, Theory, and Applications}.
\newblock United States Chess Federation, 1966.

\bibitem{minedojo}
Linxi Fan, Guanzhi Wang, Yunfan Jiang, Ajay Mandlekar, Yuncong Yang, Haoyi Zhu, Andrew Tang, De-An Huang, Yuke Zhu, and Anima Anandkumar.
\newblock Minedojo: Building open-ended embodied agents with internet-scale knowledge.
\newblock In {\em Thirty-sixth Conference on Neural Information Processing Systems Datasets and Benchmarks Track}, 2022.

\bibitem{GAN}
Ian Goodfellow, Jean Pouget-Abadie, Mehdi Mirza, Bing Xu, David Warde-Farley, Sherjil Ozair, Aaron Courville, and Yoshua Bengio.
\newblock Generative adversarial networks.
\newblock {\em Communications of the ACM}, 63(11):139--144, 2020.

\bibitem{3dgen}
Anchit Gupta, Wenhan Xiong, Yixin Nie, Ian Jones, and Barlas O{\u{g}}uz.
\newblock 3dgen: Triplane latent diffusion for textured mesh generation.
\newblock {\em arXiv preprint arXiv:2303.05371}, 2023.

\bibitem{3D-Gen-Survey3}
Negar Heidari and Alexandros Iosifidis.
\newblock Geometric deep learning for computer-aided design: A survey, 2024.

\bibitem{DDPM}
Jonathan Ho, Ajay Jain, and Pieter Abbeel.
\newblock Denoising diffusion probabilistic models.
\newblock {\em Advances in neural information processing systems}, 33:6840--6851, 2020.

\bibitem{instruct2act}
Siyuan Huang, Zhengkai Jiang, Hao Dong, Yu~Qiao, Peng Gao, and Hongsheng Li.
\newblock Instruct2act: Mapping multi-modality instructions to robotic arm actions with large language model, 2024.

\bibitem{innerm}
Wenlong Huang, Fei Xia, Ted Xiao, Harris Chan, Jacky Liang, Pete Florence, Andy Zeng, Jonathan Tompson, Igor Mordatch, Yevgen Chebotar, Pierre Sermanet, Tomas Jackson, Noah Brown, Linda Luu, Sergey Levine, Karol Hausman, and brian ichter.
\newblock Inner monologue: Embodied reasoning through planning with language models.
\newblock In {\em 6th Annual Conference on Robot Learning}, 2022.

\bibitem{holo-diffusion}
Animesh Karnewar, Andrea Vedaldi, David Novotny, and Niloy Mitra.
\newblock Holodiffusion: Training a {3D} diffusion model using {2D} images.
\newblock In {\em Proceedings of the IEEE/CVF conference on computer vision and pattern recognition}, 2023.

\bibitem{3d-gaussian}
Bernhard Kerbl, Georgios Kopanas, Thomas Leimkuehler, and George Drettakis.
\newblock 3d gaussian splatting for real-time radiance field rendering.
\newblock {\em ACM Trans. Graph.}, 42(4), jul 2023.

\bibitem{VAE}
Diederik~P Kingma and Max Welling.
\newblock Auto-encoding variational bayes.
\newblock {\em arXiv preprint arXiv:1312.6114}, 2013.

\bibitem{Free2CAD}
Changjian Li, Hao Pan, Adrien Bousseau, and Niloy~J. Mitra.
\newblock Free2cad: Parsing freehand drawings into cad commands.
\newblock {\em ACM Trans. Graph. (Proceedings of SIGGRAPH 2022)}, 41(4):93:1--93:16, 2022.

\bibitem{3D-Gen-Survey1}
Xiaoyu Li, Qi~Zhang, Di~Kang, Weihao Cheng, Yiming Gao, Jingbo Zhang, Zhihao Liang, Jing Liao, Yan-Pei Cao, and Ying Shan.
\newblock Advances in 3d generation: A survey, 2024.

\bibitem{codeaspolicy}
Jacky Liang, Wenlong Huang, Fei Xia, Peng Xu, Karol Hausman, brian ichter, Pete Florence, and Andy Zeng.
\newblock Code as policies: Language model programs for embodied control.
\newblock In {\em Workshop on Language and Robotics at CoRL 2022}, 2022.

\bibitem{Steve-1}
Shalev Lifshitz, Keiran Paster, Harris Chan, Jimmy Ba, and Sheila McIlraith.
\newblock Steve-1: A generative model for text-to-behavior in minecraft (abridged version).
\newblock In {\em NeurIPS 2023 Workshop on Goal-Conditioned Reinforcement Learning}, 2023.

\bibitem{MCU}
Haowei Lin, Zihao Wang, Jianzhu Ma, and Yitao Liang.
\newblock Mcu: A task-centric framework for open-ended agent evaluation in minecraft.
\newblock In {\em Second Agent Learning in Open-Endedness Workshop}, 2023.

\bibitem{3D-Gen-Survey2}
Jian Liu, Xiaoshui Huang, Tianyu Huang, Lu~Chen, Yuenan Hou, Shixiang Tang, Ziwei Liu, Wanli Ouyang, Wangmeng Zuo, Junjun Jiang, and Xianming Liu.
\newblock A comprehensive survey on 3d content generation, 2024.

\bibitem{FreeCAD}
J.C. Mariscal-Melgar and Pieter Hijma.
\newblock Freecad for osh automated documentation, February 2023.

\bibitem{nerf}
Ben Mildenhall, Pratul~P. Srinivasan, Matthew Tancik, Jonathan~T. Barron, Ravi Ramamoorthi, and Ren Ng.
\newblock Nerf: representing scenes as neural radiance fields for view synthesis.
\newblock {\em Commun. ACM}, 65(1):99--106, dec 2021.

\bibitem{point-e}
Alex Nichol, Heewoo Jun, Prafulla Dhariwal, Pamela Mishkin, and Mark Chen.
\newblock Point-e: A system for generating 3d point clouds from complex prompts.
\newblock {\em arXiv preprint arXiv:2212.08751}, 2022.

\bibitem{gpt-4}
OpenAI.
\newblock Gpt-4 technical report, 2023.

\bibitem{hyvin}
Shaohui Peng, Xingui Hu, Qi~Yi, Rui Zhang, Jiaming Guo, Di~Huang, Zikang Tian, Rui Chen, Zidong Du, Qi~Guo, Yunji Chen, and Ling Li.
\newblock Self-driven grounding: Large language model agents with automatical language-aligned skill learning.
\newblock {\em ArXiv}, abs/2309.01352, 2023.

\bibitem{dreamfusion}
Ben Poole, Ajay Jain, Jonathan~T. Barron, and Ben Mildenhall.
\newblock Dreamfusion: Text-to-3d using 2d diffusion.
\newblock In {\em The Eleventh International Conference on Learning Representations}, 2023.

\bibitem{mineflayer}
PrismarineJS.
\newblock Mineflayer: Create minecraft bots with a powerful, stable, and high level javascript api, also usable from python.
\newblock \url{https://github.com/PrismarineJS/mineflayer}, 2013.

\bibitem{NormalizedFlow}
Danilo Rezende and Shakir Mohamed.
\newblock Variational inference with normalizing flows.
\newblock In {\em International conference on machine learning}, pages 1530--1538. PMLR, 2015.

\bibitem{3D-GPT}
Chunyi Sun, Junlin Han, Weijian Deng, Xinlong Wang, Zishan Qin, and Stephen Gould.
\newblock 3d-{GPT}: Procedural 3d modeling with large language models, 2024.

\bibitem{lamma2}
Hugo Touvron and Louis~Martin et~al.
\newblock Llama 2: Open foundation and fine-tuned chat models, 2023.

\bibitem{Transformer}
Ashish Vaswani, Noam Shazeer, Niki Parmar, Jakob Uszkoreit, Llion Jones, Aidan~N Gomez, {\L}ukasz Kaiser, and Illia Polosukhin.
\newblock Attention is all you need.
\newblock {\em Advances in neural information processing systems}, 30, 2017.

\bibitem{Voyager}
Guanzhi Wang, Yuqi Xie, Yunfan Jiang, Ajay Mandlekar, Chaowei Xiao, Yuke Zhu, Linxi Fan, and Anima Anandkumar.
\newblock Voyager: An open-ended embodied agent with large language models.
\newblock {\em Transactions on Machine Learning Research}, 2024.

\bibitem{DEPS}
Zihao Wang, Shaofei Cai, Guanzhou Chen, Anji Liu, Xiaojian Ma, and Yitao Liang.
\newblock Describe, explain, plan and select: Interactive planning with llms enables open-world multi-task agents.
\newblock In {\em Thirty-seventh Conference on Neural Information Processing Systems}, 2023.

\bibitem{Jarvis-1}
Zihao Wang, Shaofei Cai, Anji Liu, Yonggang Jin, Jinbing Hou, Bowei Zhang, Haowei Lin, Zhaofeng He, Zilong Zheng, Yaodong Yang, Xiaojian Ma, and Yitao Liang.
\newblock Jarvis-1: Open-world multi-task agents with memory-augmented multimodal language models.
\newblock {\em arXiv preprint arXiv: 2311.05997}, 2023.

\bibitem{Hierarchical-CAD}
Xiang Xu, Pradeep~Kumar Jayaraman, Joseph~G. Lambourne, Karl~D.D. Willis, and Yasutaka Furukawa.
\newblock Hierarchical neural coding for controllable cad model generation.
\newblock In {\em Proceedings of the 40th International Conference on Machine Learning}, ICML'23. JMLR.org, 2023.

\bibitem{pku-creative-agent}
Chi Zhang, Penglin Cai, Yuhui Fu, Haoqi Yuan, and Zongqing Lu.
\newblock Creative agents: Empowering agents with imagination for creative tasks, 2023.

\bibitem{CADParser}
Shengdi Zhou, Tianyi Tang, and Bin Zhou.
\newblock Cadparser: A learning approach of sequence modeling for b-rep cad.
\newblock In Edith Elkind, editor, {\em Proceedings of the Thirty-Second International Joint Conference on Artificial Intelligence, {IJCAI-23}}, pages 1804--1812. International Joint Conferences on Artificial Intelligence Organization, 8 2023.
\newblock Main Track.

\bibitem{GITM}
Xizhou Zhu, Yuntao Chen, Hao Tian, Chenxin Tao, Weijie Su, Chenyu Yang, Gao Huang, Bin Li, Lewei Lu, Xiaogang Wang, Yu~Qiao, Zhaoxiang Zhang, and Jifeng Dai.
\newblock Ghost in the minecraft: Hierarchical agents for minecraft via large language models with text-based knowledge and memory, 2024.

\end{thebibliography}

\newpage
\appendix

\section{Broader Impacts} \label{app:broader-impacts}

In this work, we propose the Luban agent for open-ended creative building tasks in Minecraft.
Since the Minecraft game is a virtual world, the Luban agent in Minecraft will not have any positive or negative impact.
In the real world, although we demonstrated Luban's potential in performing creative building tasks on embodied robotic arms, this part is still in the prototype stage.
The future work of Luban should pay attention to the creation legality and the execution safety.

\section{Computational Resources} \label{app:computational}

Our work does not require significant local computational resources (e.g., GPU or CPU resources).
The main computational overhead of this work comes from accessing OpenAI's visual language model (i.e., the API costs of using \texttt{gpt-4-vision-preview}).
We give a rough API cost here: completing experiments of all the seeds of Luban and its ablation baseline costs about 250 USD.

\section{Implementation Details} \label{app:implementation}

\subsection{The Python CAD Modeling Library for Planning}

We simplified and encapsulated the CaDQuery project to build the Python CAD modeling library for Luban.
During the subcomponent generation phase, Luban performs modeling via the APIs in Listing \ref{app:lst:subcomponent-api}.
To create a subcomponent, LLM first initializes a panel class (specifying its x\_dim, y\_dim, and thickness).
Then, further \texttt{sub\_rect} operation can cut a rectangular hole at a certain position on the subcomponent; 
\texttt{grow\_rect} operation can extrude a rectangular boss at the specified position;
\texttt{fill\_rect} operation can fill a rectangular hole.
When modeling a subcomponent, each operation can be appended with two natural language annotations: (1) Appearance annotations, which describe the style or material of the subcomponent; (2) Verification annotations, which describe the functions that the subcomponent should have.
The appearance annotations affect the use of materials in construction actions, and the verification annotations are further used to generate actions in the pragmatic procedural verification stage.
After generating subcomponents, Luban use the APIs in Listing \ref{app:lst:assembling-api} to assemble the subcomponents by specifying the subcomponent's position and orientation.

\begin{lstlisting}[caption={The Python CAD Modeling APIs for subcomponent generation.}, label={app:lst:subcomponent-api}]
    class Panel(CADBaseObject):
    '''
    The `Panel` class is the only class you can use in building subcomponents.
    This object represents several blocks in Minecraft.
    '''
    def __init__(self, x_dim, y_dim, thickness, object_name="panel_default", annotation="", v_annotation=None):
        '''
        Method parameters: 
            x_dim: panel x_dim,
            y_dim: panel y_dim,
            thickness: panel thickness,
            object_name: object name, must be unique between different `Panel` object
            annotation: (Not Empty) Default natural language annotation for panel objects. The content of the annotation should contain the following fields: (1) Usage: e.g. window, (2) Look Like: e.g., transparent (3) Recommended blocks in Minecraft: e.g., glass.
            v_annotation: The `v_annotation` is short for verification annotation. If the building result of the subcomponent's building operation (subcomponent initialization is also considered a build operation) needs to meet some functional requirements, please indicate it in `v_annotation`. The `v_annotation` is usually a Python dict with the following contents: v_annotation = {
	    'type': "xxx", # The type of the v_annotation, only support 2 types: "planner" or "env". The "env" type indicates that the verifying the functional features of the subcomponent must interact with the Minecraft Game environment through an agent to determine whether it is passed. The "planner" type means that verifying the functional features of the subcomponent can be done only during the planning phase, without interaction with the environment.
	    'anno': "xxx" # The content of the v_annotation. When the "type" is "env", the content of "anno" should be about how to interact with this subcomponent in the Minecraft game environment. When the "type" is "planner", the content of "anno" should be about what constraints the subcomponent should satisfy during the planning phase.
}
	
            return value: None
        When initializing a `Panel` Class, you get a general rectangle panel (x_dim, y_dim, thickness) with default center position at (0, 0, thickness / 2) (Unchangeable). 
        You can use other methods in this class to perform further operations on the general rectangle panel.
        You can get the name of the Panel object with `self.name`.
        '''
    def sub_rect(self, pos, rect_shape, sub_rect_name, v_annotation=None):
        '''
        Method parameters: 
            pos: The center position of the rectangle hole (offset relative to the xy-center, i.e. (0, 0), of the panel object), a tuple with 2 elements (x, y).
            rect_shape: The size of the rectangle hole, a tuple with 2 elements (x_dim, y_dim).
            sub_rect_name: The name of the rectangle hole, str. The name must be unique within the current `Panel` object but can be repeated across different `Panel` objects.
            v_annotation: The `v_annotation` is short for verification annotation. If the building result of the subcomponent's building operation (subcomponent initialization is also considered a build operation) needs to meet some functional requirements, please indicate it in `v_annotation`. The `v_annotation` is usually a Python dict with the following contents: v_annotation = {
	    'type': "xxx", # The type of the v_annotation, only support 2 types: "planner" or "env". The "env" type indicates that the verifying the functional features of the subcomponent must interact with the Minecraft Game environment through an agent to determine whether it is passed. The "planner" type means that verifying the functional features of the subcomponent can be done only during the planning phase, without interaction with the environment.
	    'anno': "xxx" # The content of the v_annotation. When the "type" is "env", the content of "anno" should be about how to interact with this subcomponent in the Minecraft game environment. When the "type" is "planner", the content of "anno" should be about what constraints the subcomponent should satisfy during the planning phase.
}

            return value: None
    
        Dig a rectangle hole on the panel, with reference name `sub_rect_name.` 
        When using the `sub_rect` method, you need to determine the appropriate value for the `pos` parameter based on the xy-center position of the `Panel` object.
        You can refer to the rect hole by `sub_rect_name` in other operations.
        '''
    
    def fill_rect(self, rect_hole_name, fill_name, annotation="", v_annotation=None):
        '''
        Method parameters: 
            rect_hole_name: The name of a rectangle hole. Please make sure the name exists (already created by another operation)
            fill_name: The name of the window, str. The name must be unique within the current `Panel` object but can be repeated across different `Panel` objects.
            annotation: Natural language annotation for the `fill_rect` operation generated blocks. The content of the annotation should contain the following fields: (1) Usage: e.g. window, (2) Look Like: e.g., transparent (3) Recommended blocks in Minecraft: e.g., glass.
            v_annotation: The `v_annotation` is short for verification annotation. If the building result of the subcomponent's building operation (subcomponent initialization is also considered a build operation) needs to meet some functional requirements, please indicate it in `v_annotation`. The `v_annotation` is usually a Python dict with the following contents: v_annotation = {
    'type': "xxx", # The type of the v_annotation, only support 2 types: "planner" or "env". The "env" type indicates that the verifying the functional features of the subcomponent must interact with the Minecraft Game environment through an agent to determine whether it is passed. The "planner" type means that verifying the functional features of the subcomponent can be done only during the planning phase, without interaction with the environment.
    'anno': "xxx" # The content of the v_annotation. When the "type" is "env", the content of "anno" should be about how to interact with this subcomponent in the Minecraft game environment. When the "type" is "planner", the content of "anno" should be about what constraints the subcomponent should satisfy during the planning phase.
}
    
            return value: None
    
        Fill a rectangle hole with blocks specified by `annotation`, with reference name `fill_name.` 
        You do not need to create a `Panel` object representing the window or door, just call this method (See Example 3).
        You can refer to the filled result by `fill_name` in other operations, but the filled result is generally not referenced.
        '''


     
     def grow_rect(self, pos, rect_shape, thickness, grow_rect_name, base_rect_name, annotation="", v_annotation=None):
         '''
         Method parameters: 
             pos: The center position of the rectangle column (offset relative to the center of the panel object), a tuple with 2 elements (x, y).
             rect_shape: The size of the rectangle hole, a tuple with 2 elements (x_dim, y_dim).
             thickness: The thickness of the rectangle column.
             grow_rect_name: The name of the rectangle column. The name must be unique within the current `Panel` object but can be repeated across different `Panel` objects.
             base_rect_name: The name of the window, str. Please make sure the name exists (already created by another operation).
             annotation: Natural language annotation for the `grow_rect` operation. The content of the annotation should contain the following fields: (1) Usage: e.g. window, (2) Look Like: e.g., transparent (3) Recommended blocks in Minecraft: e.g., glass.
             v_annotation: The `v_annotation` is short for verification annotation. If the building result of the subcomponent's building operation (subcomponent initialization is also considered a build operation) needs to meet some functional requirements, please indicate it in `v_annotation`. The `v_annotation` is usually a Python dict with the following contents: v_annotation = {
    'type': "xxx", # The type of the v_annotation, only support 2 types: "planner" or "env". The "env" type indicates that the verifying the functional features of the subcomponent must interact with the Minecraft Game environment through an agent to determine whether it is passed. The "planner" type means that verifying the functional features of the subcomponent can be done only during the planning phase, without interaction with the environment.
    'anno': "xxx" # The content of the v_annotation. When the "type" is "env", the content of "anno" should be about how to interact with this subcomponent in the Minecraft game environment. When the "type" is "planner", the content of "anno" should be about what constraints the subcomponent should satisfy during the planning phase.
}
            
            return value: None
    
        Grow/extrude a rectangle column with blocks specified by `annotation`, along the thickness axis (z+ direction).
        The rectangle column (with reference name `grow_rect_name`) starts from the z-axis (i.e., thickness axis) height of the `base_rect_name`, and extends `thickness` along the z-axis.
        For example, when `base_rect_name == self.name`, the rectangle column starts with `z = self.thickness`, and ends with `z = self.thickness + thickness`.
        This method is often used to build pyramid-like objects such as roofs, by calling this method hierarchically.
        More details and illustrations of this method can be found in the code and comments of example 4 and 5.
        
        You can refer to the rectangle column by `grow_rect_name` in other operations.
        '''

\end{lstlisting}

\begin{lstlisting}[caption={The Python CAD Modeling APIs for assembling.}, label={app:lst:assembling-api}]
class CADAssembly(object):
    def __init__(self):
        '''
        Method parameter: 
            None

        Return value: 
            None
        
        No parameters are required when creating a new `CADAssembly` object. 
        A series of operations on the CADAssembly object can complete the assembly of sub-components.
        '''
    
    def add_object(self, obj):
        '''
        Method parameter: 
            obj: The subcomponent object to be added to the assembly.

        Return value: 
            None
        
        This method is used to add a new object to the CADAssembly (no return value). Each object can only be added once.
        '''

    
    def set_object_pos(self, obj, pos):
        '''
        Method parameter: 
            obj: The object whose position is to be set. Please make sure the object has been added to class CADAssembly before calling this method on the object.
            pos: Target position. A tuple containing 3 floating point or integer values.
        
        Return value: 
            None
        
        This method can set the position `pos` of the reference point of the subcomponent `obj`. 
        '''


    def set_object_orientation(self, obj, orientation):
        '''
        Method parameter: 
            obj: The object whose orientation is to be set. Please make sure the object has been added to class CADAssembly before calling this method on the object.
            orientation: 
                Target orientation, str.
                There are 6 legal values, namely `north`, `south`, `east`, `west`, `up` and `down`, which represent setting the orientation of `obj` to the corresponding `orientation`. 
        
        Return value: 
            None
        
        This method can set the orientation `orientation` of the object `obj`.
        '''
\end{lstlisting}

\section{Benchmark Details}  \label{app:benchmark}

\subsection{Comparisons with Other Benchmarks Involving Creative Tasks} \label{app:baseline-compare}
The concept of creative tasks has also been introduced in other work on Minecraft benchmarks, including MineDojo \cite{minedojo}, Minecraft SkillForge \cite{Groot}, and Creative Agent \cite{pku-creative-agent}.
Our work focuses on building open-ended creative building agents that can fill the gap between abstract criteria and concrete verification actions via autonomous embodied verification.
From this perspective, none of the above three benchmarks are applicable to this task and the reasons are listed as follows:
(1) The Minedojo benchmark consists of creative exploration of the Minecraft world, which is inconsistent with the focus of this work: building.
(2) The Minecraft SkillForge benchmark creation tasks are very simple, involve only a few blocks ($<10$), and do not have any criteria regarding functionality, so it is not suitable for this work.
(3) The creative tasks in the benchmark of Creative agent work are of comparable complexity to our work but lack functional criteria.

\subsection{Task Illustrations}

\begin{figure}[h]
    \centering
    \includegraphics[width=\linewidth]{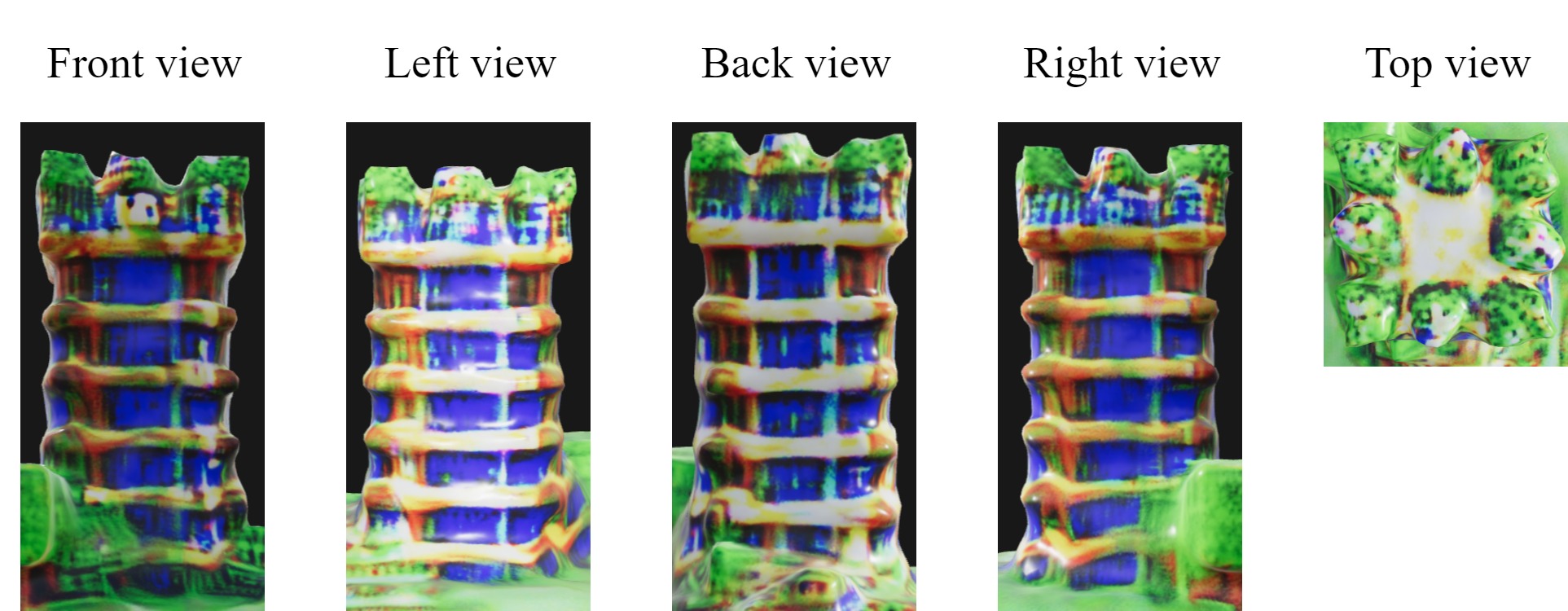}
    \caption{The multi-view images part of the \texttt{arrow-tower} task.}
    \label{app:fig:arrow}
\end{figure}

There are 5 tasks in the benchmark.
Each task's instructions consist of two parts:
(1) Text contains the building description, the functional requirements, and some building suggestions.
(2) Multi-view images of example buildings that match the building description, generated by a generic text-to-3D service (we use \texttt{meshy.ai} in this work).
The 5 task instructions in the benchmark are listed below:

1. Task \texttt{arrow-tower}, text (Listing \ref{app:lst:arrow}), image (Figure \ref{app:fig:arrow}).

\begin{lstlisting}[caption={The text part of the \texttt{arrow-tower} task.}, label={app:lst:arrow}]
Building: A simple solid arrow tower similar to the tower in images. The arrow tower has an external ladder to the top of the tower.
    
Building Suggestions: 
    You should build the arrow tower by stacking layer by layer. 
    The whole ladder and tower must above the ground plane. 
    Do not generate multiple subcomponents representing the ladder, you should merge these subcomponents into one subcomponent.
    
Functional Requirements: I can climb to the top of the tower through the ladder external. The arrow tower must slightly higher than the column in Minecraft Game I gave.
    
Key Design Parameters: The height of the arrow tower. The height of the arrow-tower depends on the height of the column. If the arrow tower does not higher than the column, try increasing the height of the arrow tower.
\end{lstlisting}
    
\begin{figure}[h]
    \centering
    \includegraphics[width=\linewidth]{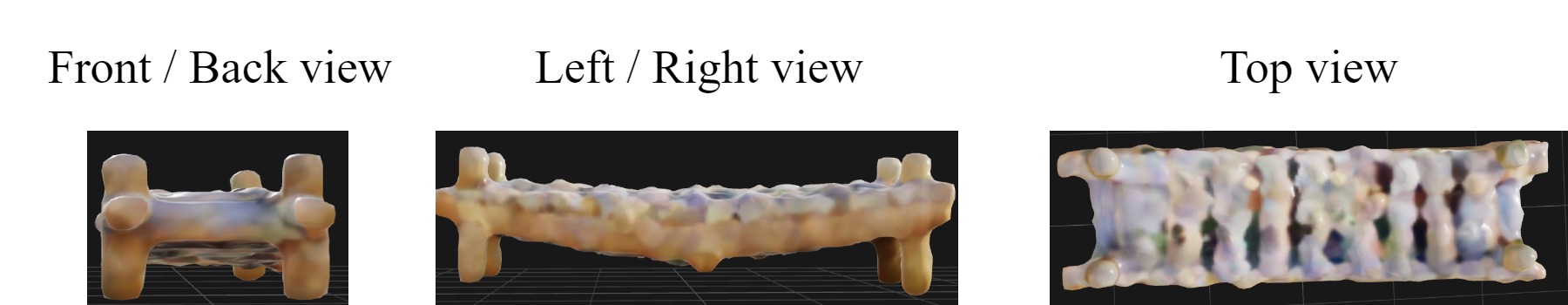}
    \caption{The multi-view images part of the \texttt{bridge} task.}
    \label{app:fig:bridge}
\end{figure}

2. Task \texttt{bridge}, text (Listing \ref{app:lst:bridge}), image (Figure \ref{app:fig:bridge}).

\begin{lstlisting}[caption={The text part of the \texttt{bridge} task.}, label={app:lst:bridge}]
Building: A simple east-west plank bridge similar to the bridge in images.

Building Suggestions: Do not create any subcomponents representing the bridge's support and omit unnecessary decorative subcomponents and elements. Assume the water surface of the river is the level of the ground.
    
Functional Requirements: The bridge should connect the east and west banks of the river for players to cross.
    
Key Design Parameters: The length of bridge. The length of the bridge depends on the width of the river, if the bridge cannot span the river, try increasing the length of the bridge.
\end{lstlisting}
        
\begin{figure}[h]
    \centering
    \includegraphics[width=\linewidth]{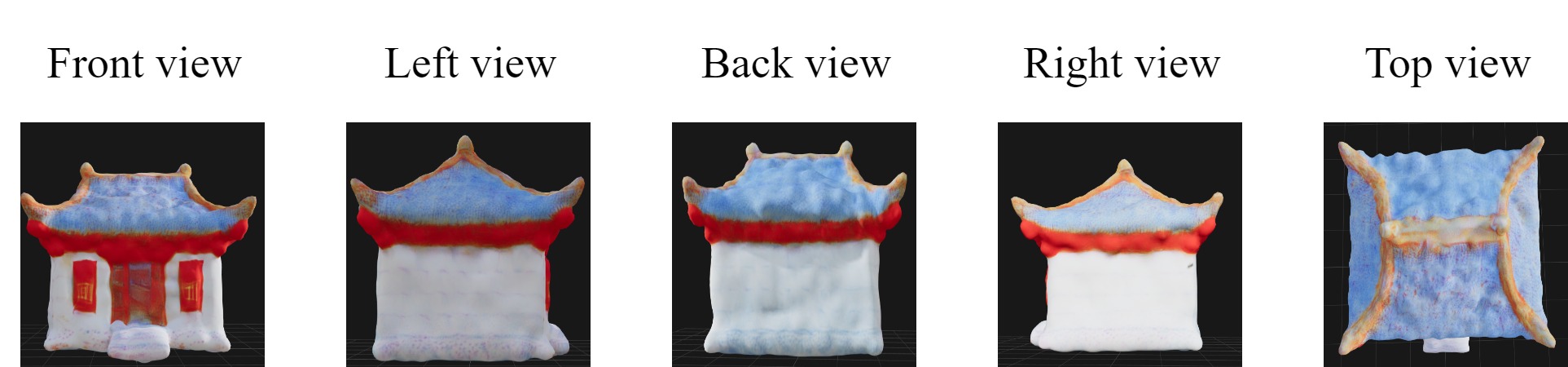}
    \caption{The multi-view images part of the \texttt{chinese-ancient-house} task.}
    \label{app:fig:chinese-ancient-house}
\end{figure}

3. Task \texttt{chinese-ancient-house}, text (Listing \ref{app:lst:chinese-ancient-house}), image (Figure \ref{app:fig:chinese-ancient-house}).

\begin{lstlisting}[caption={The text part of the \texttt{chinese-ancient-house} task.}, label={app:lst:chinese-ancient-house}]
Building: A simple Chinese ancient house similar to the house in images.

Building Suggestions: Do not create any subcomponents representing the house's foundation or interior floor. The house should not be larger than 10x10 blocks.
    
Functional Requirements: This house should have a door through which the player can enter the interior of the house.

Key Design Parameters: The size and position of the door. If the player cannot enter the house, try adjusting the size and position of the door.    
\end{lstlisting}
            
\begin{figure}[h]
    \centering
    \includegraphics[width=\linewidth]{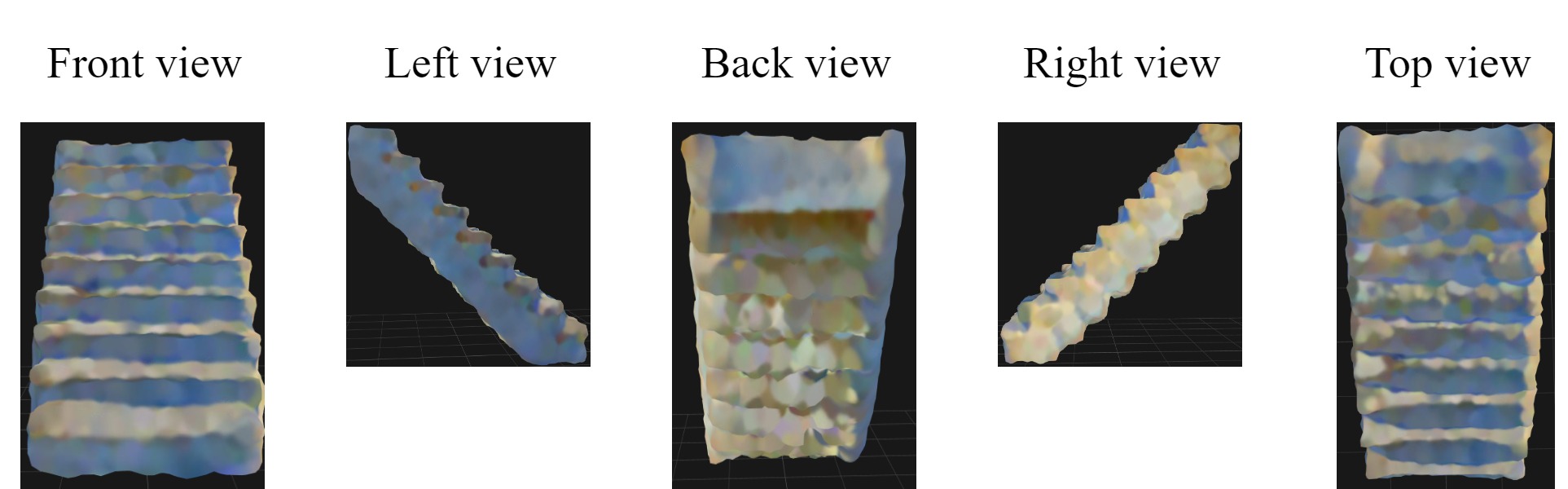}
    \caption{The multi-view images part of the \texttt{stair} task.}
    \label{app:fig:stair}
\end{figure}

4. Task \texttt{stair}, text (Listing \ref{app:lst:stair}), image (Figure \ref{app:fig:stair}).
\begin{lstlisting}[caption={The text part of the \texttt{stair} task.}, label={app:lst:stair}]
Building: A simple stair similar to the stair in images.

Building Suggestions: You can build it layer by layer.
    
Functional Requirements: The stair should high enough for players to climb the cliff.
    
Key Design Parameters: The height of the stair. The height of the stair depends on the height of the cliff. If the stair cannot help the players climb the cliff , try increasing the height of the stair.
\end{lstlisting}
                
\begin{figure}[h]
    \centering
    \includegraphics[width=\linewidth]{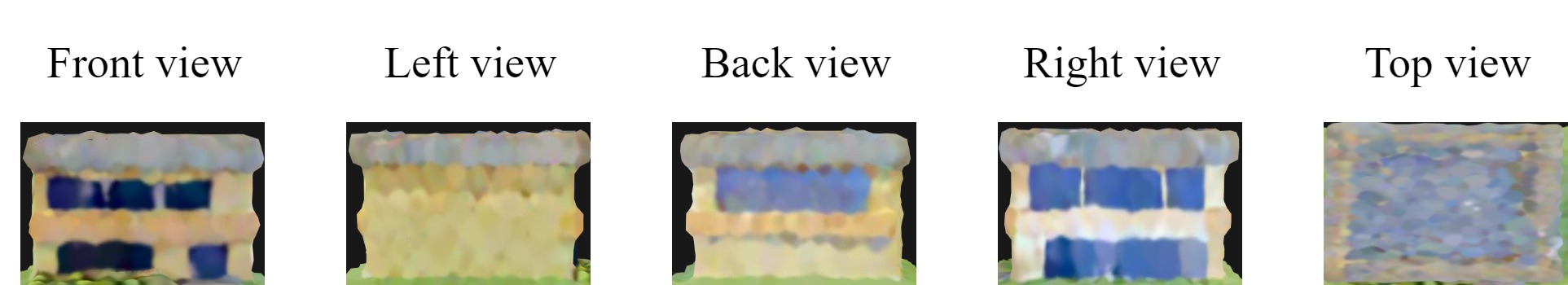}
    \caption{The multi-view images part of the \texttt{two-story-house} task.}
    \label{app:fig:two-story-house}
\end{figure}

5. Task \texttt{two-story-house}, text (Listing \ref{app:lst:two-story-house}), image (Figure \ref{app:fig:two-story-house}).
\begin{lstlisting}[caption={The text part of the \texttt{two-story-house} task.}, label={app:lst:two-story-house}]
Building: A simple two story house (the ground floor and the first floor) with flat roof similar to the house in images.

Building Suggestions: Do not create any subcomponents representing the house's foundation or interior floor. The house should not be larger than (wide x length) 10x10 blocks.
    
Functional Requirements: This house should have a door through which the player can enter interior of the house's ground floor.
    
Key Design Parameters: The size and position of the door. If the player cannot enter interior of the house's ground floor, try adjusting the size and position of the door.
\end{lstlisting}
                    
\section{Additional Experiment Results} \label{app:add-exp}

\subsection{Showcases} \label{app:add-exp-show}

\begin{figure}
    \centering
    \includegraphics[width=\linewidth]{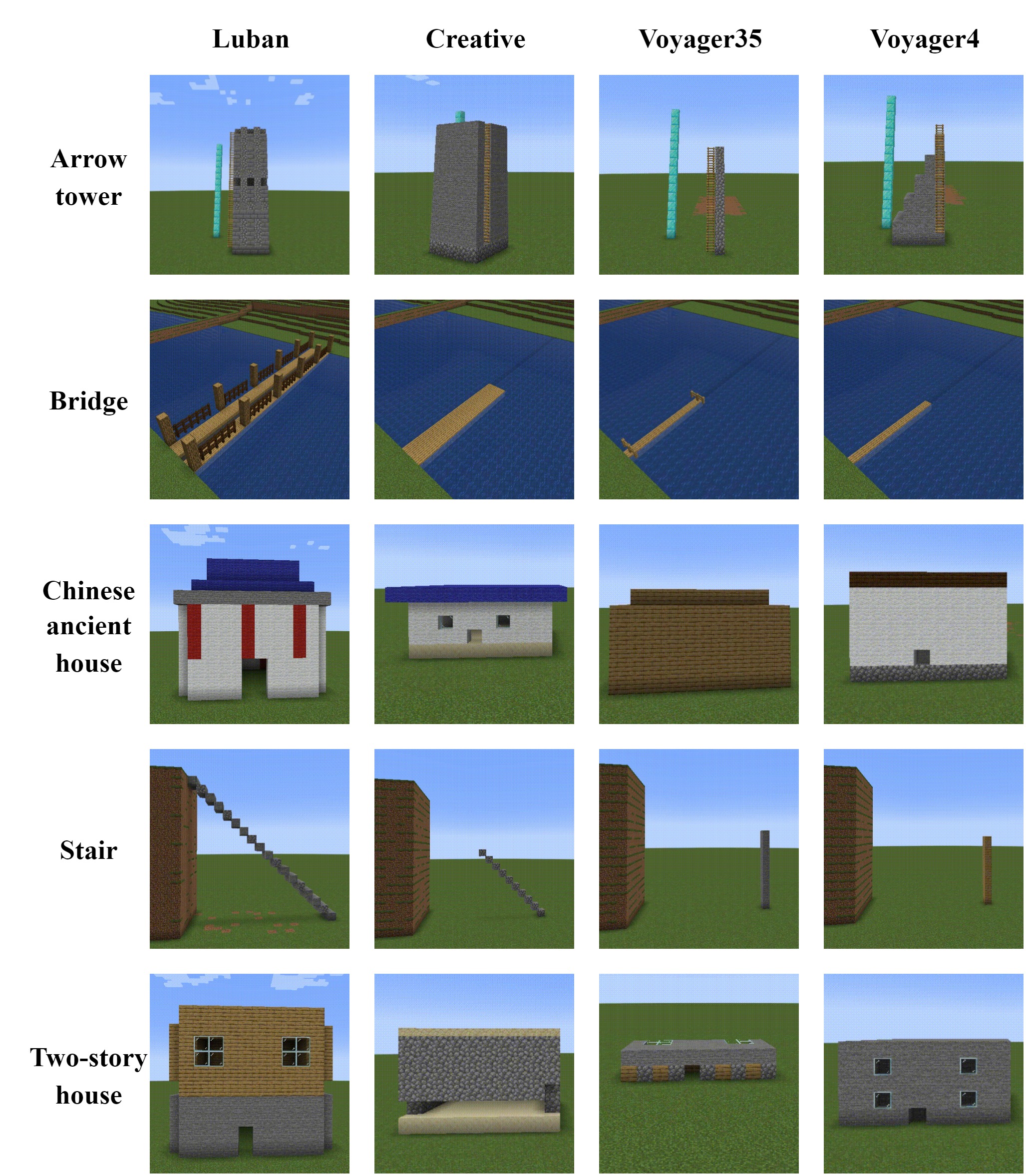}
    \caption{The showcases (from one of the three seeds) of Luban vs other baselines.}
    \label{app:fig:showcase-vs-other}
\end{figure}

\begin{figure}
    \centering
    \includegraphics[width=\linewidth]{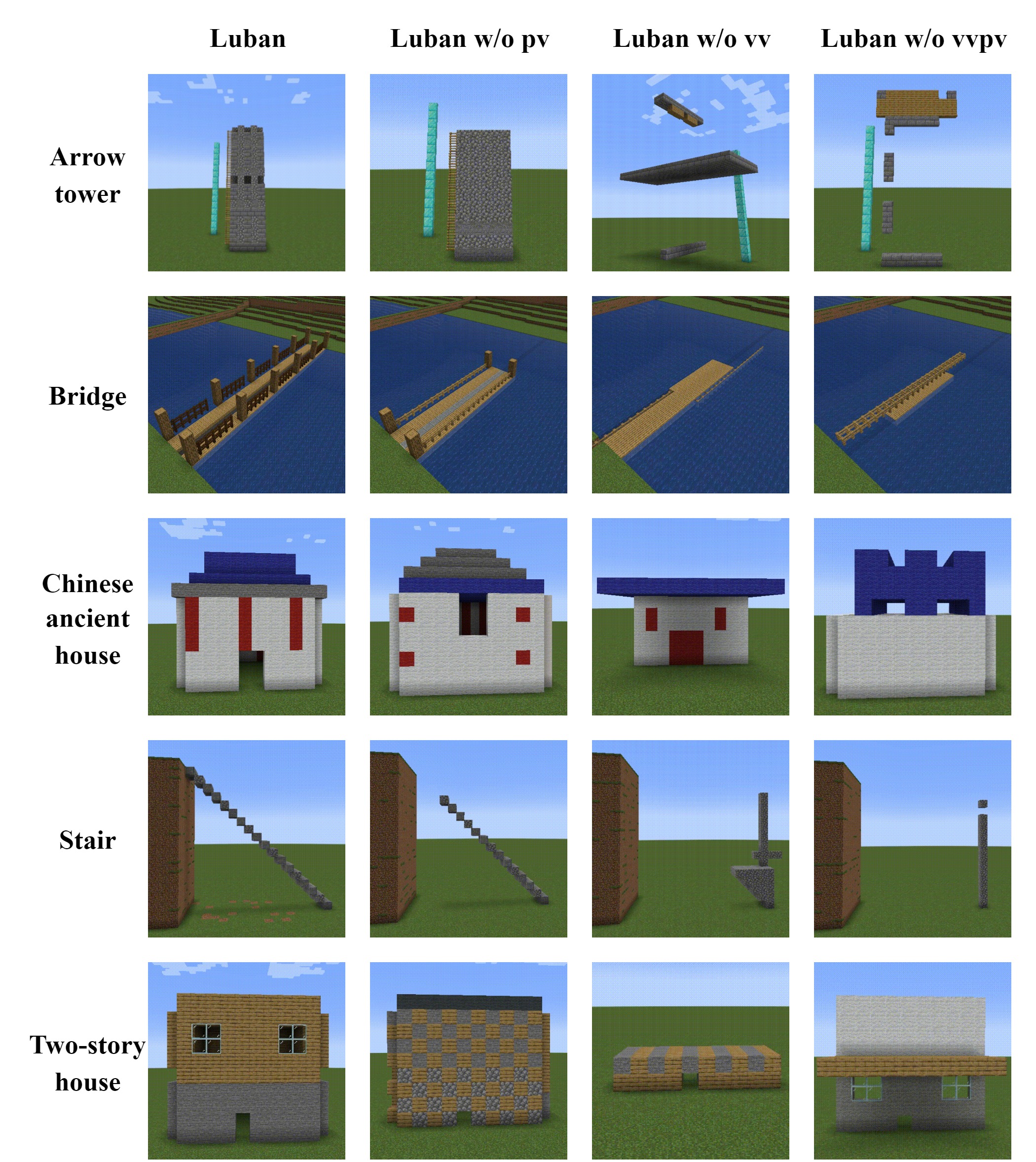}
    \caption{The showcases (from one of the three seeds) of Luban vs ablation baselines.}
    \label{app:fig:showcase-vs-ablation}
\end{figure}

We present showcases of Luban v.s. other baselines:
(1) Figure \ref{app:fig:showcase-vs-other} shows the building result comparisons between Luban and other method baselines
(2) Figure \ref{app:fig:showcase-vs-ablation} shows the building result comparisons between Luban and ablation baselines.

\subsection{Case Study} \label{app:add-exp-case}

\begin{figure}
    \centering
    \includegraphics[width=\linewidth]{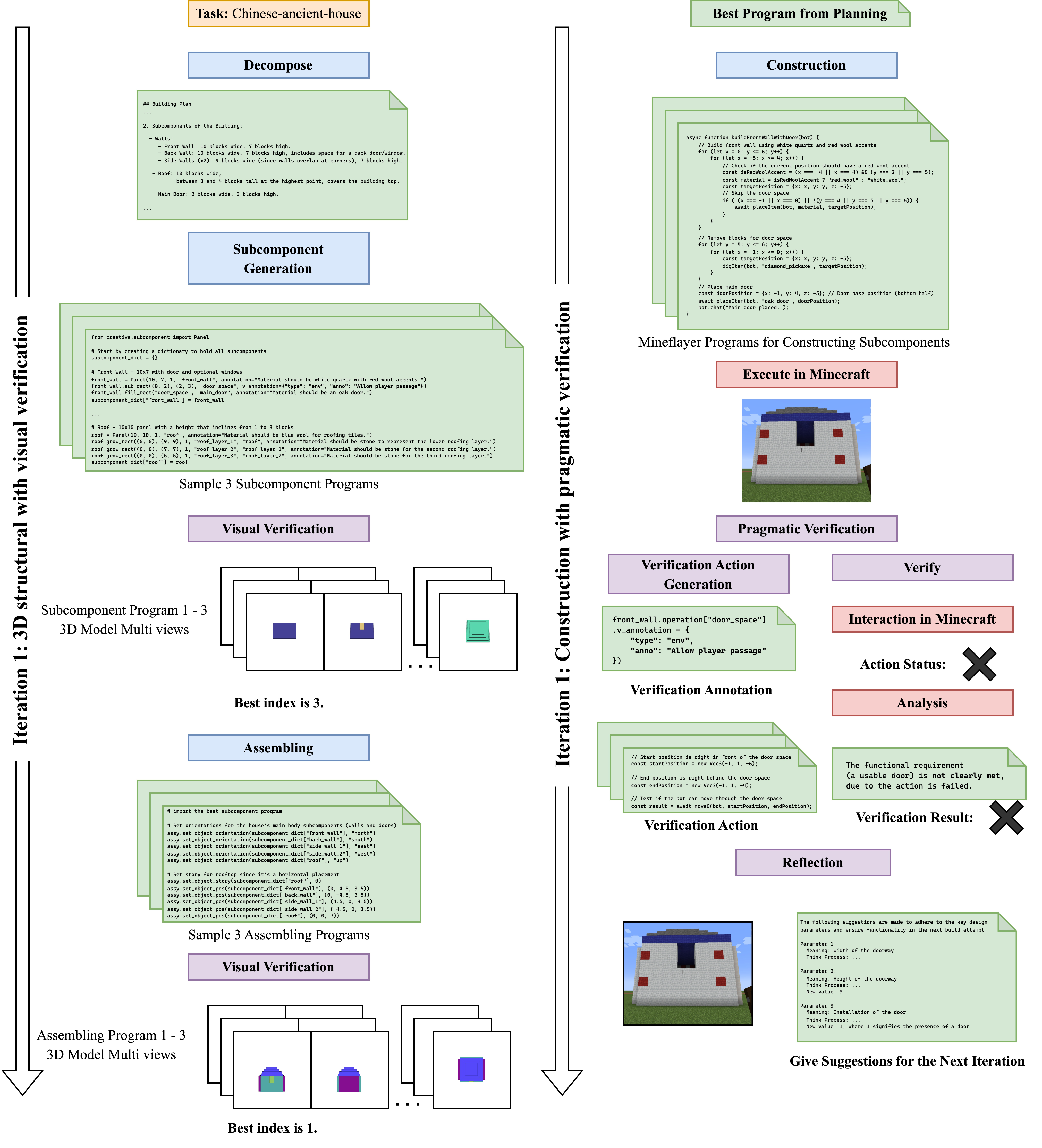}
    \caption{The case study of the \texttt{chinese-ancient-house} task (the first iteration).}
    \label{app:fig:case-study-iter-1}
\end{figure}

\begin{figure}
    \centering
    \includegraphics[width=\linewidth]{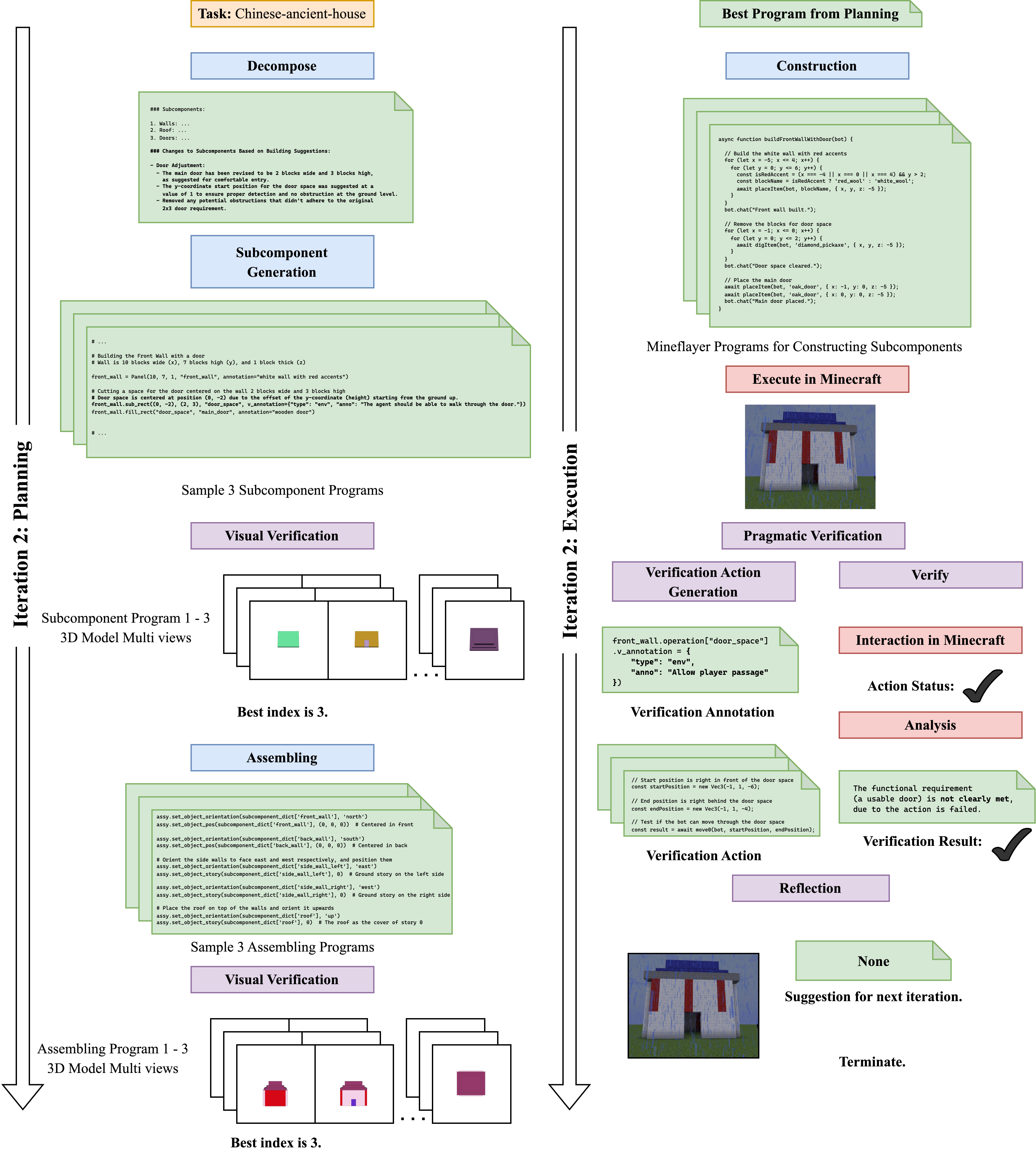}
    \caption{The case study of the \texttt{chinese-ancient-house} task (the second iteration).}
    \label{app:fig:case-study-iter-2}
\end{figure}

We present a case study of Luban when completing the \texttt{chinese-ancient-house} task, which takes two iterations (as shown in Figure \ref{app:fig:case-study-iter-1} and Figure \ref{app:fig:case-study-iter-2}).
In the first iteration, Luban plans and builds a house. 
However, the door of the house does not start from the ground, so the door is not pragmatic in the environment.
Luban discovers this error through pragmatic verification, whose autonomous verification actions are proposed by Luban based on the task instructions. 
After reflection, Luban successfully fixes the error and builds a pragmatic house in the second iteration.

\newpage
\section{Human Study} \label{app:human-study}

\subsection{Participants}
We recruited 15 human evaluators to evaluate Minecraft creative build results.
The Minecraft gaming experience of these human evaluators ranges from `never played at all' to `playing time $\ge$ 20 hours', and the distribution is shown in Table \ref{app:table:human}.
Each evaluator is asked to conduct two evaluations on all five tasks' results: (1) multi-dimensional quality ratings; (2) one-to-one comparison.
Each evaluator spent a total of 80 to 120 minutes on the two evaluations.
We paid an average of $\sim$ 8.80 USD per hour - a standard human evaluator's wage in our region.

\begin{table}
    \centering
    \caption{
        Minecraft experience statistics table for human evaluators.
    }
    \label{app:table:human}
    \begin{tabular}{c|cccc}
        \toprule
        Game Hours & \texttt{never} & $(0, 5]$ & $(5, 20]$ & $\ge 20$ \\
        \midrule
        Count & 1 & 3 & 2 & 9 \\
        \bottomrule
    \end{tabular} 
\end{table}

\subsection{Questionnaire and Interface}

\begin{figure}[htbp]
    \centering
    \includegraphics[width=\linewidth]{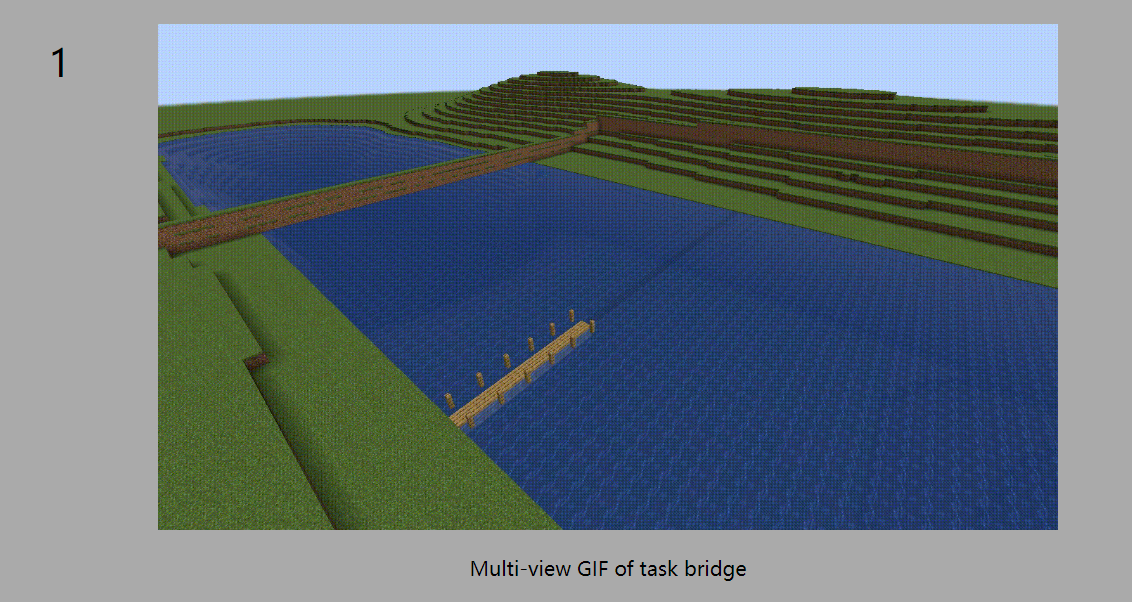}
    \caption{The local web page presents videos for multi-dimensional quality ratings (take the \texttt{bridge} task as an example)}
    \label{app:fig:h1-web}
\end{figure}

\textbf{Multi-dimensional quality ratings.}
This part requires human evaluators to perform multi-dimensional rating on $7\times 5 \times 3 = 105$ (7 baselines, 5 tasks, and 3 seeds per baseline) 8-second multi-view videos of the building.
Among the tasks in the benchmark, three tasks (i.e., \texttt{arrow-tower}, \texttt{bridge}, and \texttt{stair}) have 5-dimensional ratings, and two tasks (\texttt{chinese-ancient-house} and \texttt{two-story-house}) have 4-dimensional ratings (without functional enviromnent), and the score range of each dimension ranges from 1 to 5.
The questionnaires were grouped by task, and results from different baselines were anonymously shuffled.
When conducting the evalution, each evaluator was presented with 3 materials: 
(1) the current task's instruction (Please refer more details in Appendix \ref{app:benchmark}); (2) a questionnaire to collect ratings (Listing \ref{app:lst:h1-q}); (3) a local web page showing seeds from all baselines of the current task (Figure \ref{app:fig:h1-web}).

\begin{lstlisting}[caption={The questionnaire to collect ratings (take the \texttt{bridge} task as an example). The descriptions of two functional correctness items in the questionnaire changed depending on the task.}, label={app:lst:h1-q}]
The questionnaire for task bridge.

1. Ratings of result (1).
    
    - Appearance: The extent to which the building conforms to the semantics and appearance of the text and multi-view images descriptions in the task instructions. Your rating here: <TODO, answering an integer value ranges 1 to 5>.
    
    - Complexity: The complexity of the building (e.g. structural / design details). Your rating here: <TODO, answering an integer value ranges 1 to 5>.
    
    - Aesthetics: The extent to which the building is aesthetic. Your rating here: <TODO, answering an integer value ranges 1 to 5>.
    
    - Functional building-level: To what extent does this building meet the functional requirements building-level (i.e., The bridge deck is walkable for players and the handrails prevent players from falling). Your rating here: <TODO, answering an integer value ranges 1 to 5>.
    
    - Functional enviromnent-level: To what extent does this building meet the functional requirements enviromnent-level (i.e., The bridge allows players across the river). Your rating here: <TODO, answering an integer value ranges 1 to 5>.


2. Ratings of result (2).

    ...
\end{lstlisting}

\begin{figure}[htbp]
    \centering
    \includegraphics[width=\linewidth]{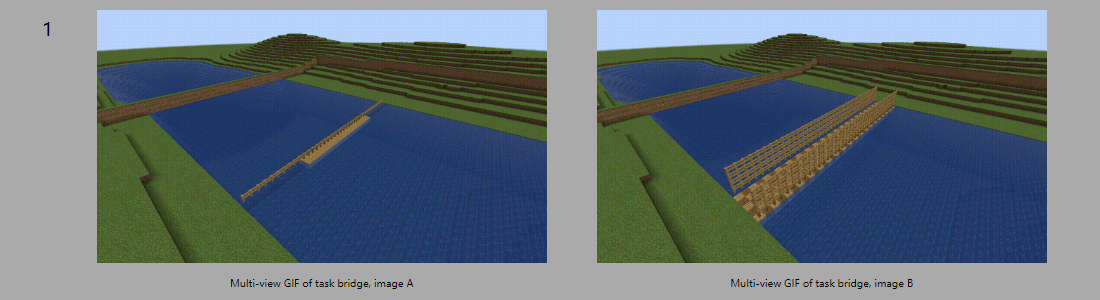}
    \caption{The local web page for presenting one-to-one comparison video pairs (take the \texttt{bridge} task as an example)}
    \label{app:fig:h23-web}
\end{figure}

\textbf{One-to-one comparison.}
This part requires human evaluators to perform one-to-one comparison of the building results. 
Considering the high labor cost brought by pair-wise comparison, we split the one-to-one comparison into two parts: (1) comparisons between Luban and other method baselines (involving 4 baselines); (2) comparisons between Luban and ablation baselines (involving 4 baselines).
Under this splitting setup, each part still requires $5\times \binom{4}{2} \times 3^2 = 270$ (5 tasks, 2 out of 4 baselines are selected, and $(\texttt{num\_seed}=3)^2$ different matches) comparisons.
To further reduce labor costs, we randomly sample $k = 3$ mathces in any two baselines, so there are a total of $5\times \binom{4}{2} \times k = 30k = 90$ comparisons in each part.
The two part questionnaires were also grouped by task, and results from different baselines were anonymously shuffled.
When conducting the evalution, each evaluator was presented with 3 materials: 
(1) the current task's instruction (Please refer more details in Appendix \ref{app:benchmark}); (2) a questionnaire to collect the winner (Listing \ref{app:lst:h23-q}); (3) a local web page showing each one-to-one comparison of the current task (Figure \ref{app:fig:h23-web}).

\begin{lstlisting}[caption={The questionnaire to collect one-to-one comparison winners (take the \texttt{bridge} task as an example).}, label={app:lst:h23-q}]
The questionnaire for task bridge.
    
1. The winner of the comparison (1).
        
    - Based on the instruction of the bridge task,  which one (A or B) is better overall? Your answer here: <TODO, answering A or B>.
        
2. The winner of the comparison (2).
    
    ...
\end{lstlisting}

\end{document}